\pdfoutput=1

\documentclass[11pt]{article}

\usepackage[final]{scope}

\usepackage{times}
\usepackage{latexsym}

\usepackage[T1]{fontenc}

\usepackage[utf8]{inputenc}

\usepackage{microtype}

\usepackage{inconsolata}

\usepackage{kotex}
\usepackage{subcaption}
\usepackage{booktabs}
\usepackage{multirow}
\usepackage{makecell}
\usepackage{graphics}
\usepackage{amssymb}
\usepackage{bbm}
\usepackage{amsmath}
\usepackage{lipsum}
\usepackage
{graphicx}%
\usepackage[english]{babel}
\usepackage{algorithm}
\usepackage[noend]{algpseudocode}
\usepackage{enumitem}
\makeatletter
\algnewcommand{\LineComment}[1]{\Statex \hskip\ALG@thistlm \(\triangleright\) #1}
\makeatother

\usepackage{xcolor}
\usepackage{multicol}

\usepackage{etoolbox}

\title{Controlled Text Generation for Black-box Language Models via\\Score-based Progressive Editor}

\author{Sangwon Yu$^1$\hspace{5mm} Changmin Lee$^2$\hspace{5mm} Hojin Lee$^2$\hspace{5mm} Sungroh Yoon$^{1,3,}$\Thanks{\ Corresponding author} \\
   $^1$Department of Electrical and Computer Engineering, Seoul National University\\
   $^2$Kakao Corp. \\
   $^3$Interdisciplinary Program in Artificial Intelligence, Seoul National University \\   
   {\tt\normalsize\{dbtkddnjs96, sryoon\}@snu.ac.kr}\hspace{6mm}
   {\tt\normalsize\{louie.m, lambda.xprime\}@kakaocorp.com}}

\begin{document}
\maketitle
\begin{abstract}
Controlled text generation is very important for the practical use of language models because it ensures that the produced text includes only the desired attributes from a specific domain or dataset.
Existing methods, however, are inapplicable to black-box models or suffer a significant trade-off between controlling the generated text and maintaining its fluency.
This paper introduces the Score-based Progressive Editor (ScoPE), a novel approach designed to overcome these issues. 
ScoPE modifies the context at the token level during the generation process of a backbone language model. 
This modification guides the subsequent text to naturally include the target attributes. 
To facilitate this process, ScoPE employs a training objective that maximizes a target score, thoroughly considering both the ability to guide the text and its fluency.
Experimental results on diverse controlled generation tasks demonstrate that ScoPE can effectively regulate the attributes of the generated text while fully utilizing the capability of the backbone large language models.
Our codes are available at \url{https://github.com/ysw1021/ScoPE}.
\end{abstract}

\section{Introduction}
Modern language models can now generate text as fluently as humans in response to any given sequence or instruction
(\citealp{Radford2018ImprovingLU_gpt1}; \citealp{radford2019language_gpt2}; \citealp{Brown2020-of_gpt3}; \citealp{Ouyang2022-nv_instructgpt}; \citealp{Zhang2022-zo_opt}; \citealp{Touvron2023-fp_llama}; \citealp{llama2}; \citealp{jiang2023mistral}). However, the generated text may carry potential risks, such as harmful expressions or inappropriate content (\citealp{gehman-etal-2020-realtoxicityprompts}; \citealp{liu-etal-2021-dexperts}; \citealp{Lu2022-gw_quark}). 
If the previous context includes non-preferred attributes, these attributes can potentially appear in the generated text.
Therefore, controlled text generation, which aims to generate text constrained to target domain attributes regardless of the given context, is crucial for addressing the current issues in language models (\citealp{Dathathri2019-tf_pplm}; \citealp{Khalifa2020-mo_dpg}; \citealp{Qian2022-tq_prefixcont}; \citealp{Meng2022-wl_nado}; \citealp{Qin2022-he_cold}; \citealp{Ma2023-bnprefixcont2}). 

Recently, many large language models, especially those exceeding hundreds of billion parameters, are presented as de facto black-box models with limited access to model parameters  (\citealp{OpenAI2022}; \citealp{OpenAI2023-ll_gpt4}; \citealp{team2023gemini}; \citealp{anthropic2024claude}). 
Consequently, most existing approaches to controlled text generation that require access to the model's parameters are either inapplicable or have limitations in these black-box scenarios.
While access to the language model parameters is unavailable, some previous works assume access to the output token distribution and have studied controlled generation by manipulating this distribution without tuning the parameters (\citealp{Krause2020-xi_gedi}; \citealp{Yang2021-rr_fudge}; \citealp{Arora2022-lf_director}).
However, this approach significantly diminishes the fluency of the generated text, as it manipulates the output distribution based on the previous context. Therefore, there is a requirement for a novel approach to effectively leverage the generation performance of black-box large language models.

\begin{figure*}[t]
     \centering
     \includegraphics[width=\textwidth]{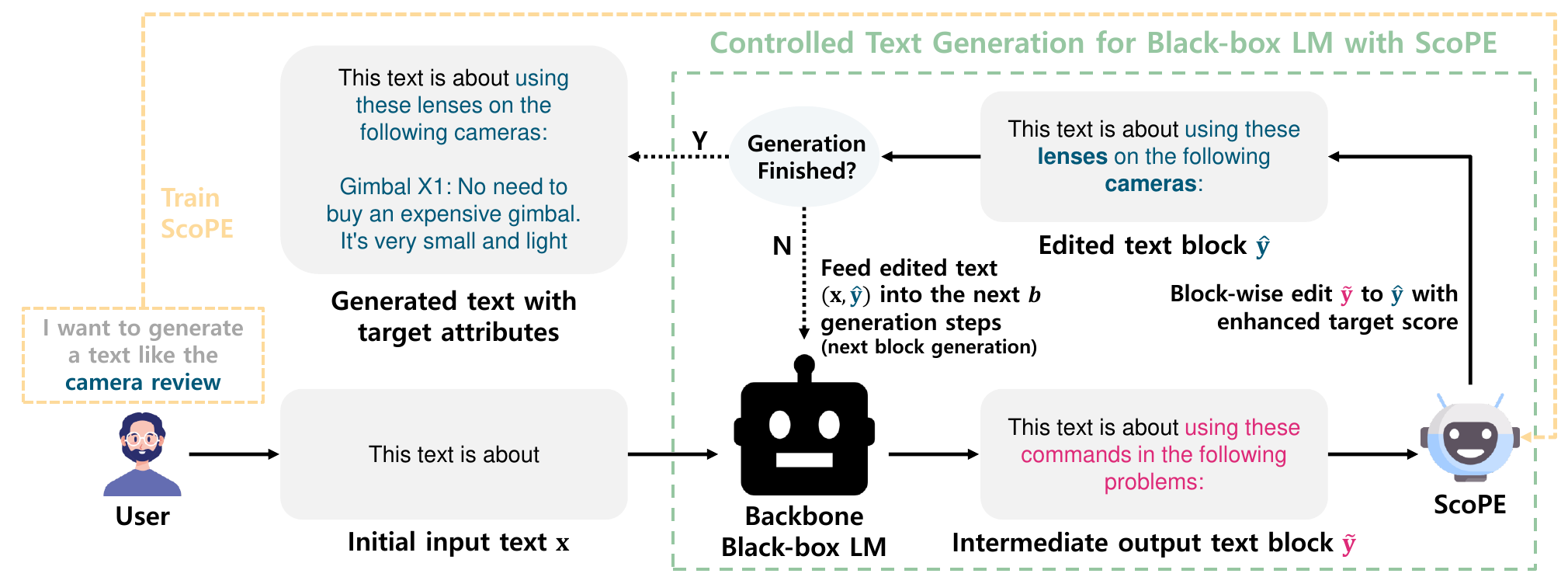}
     \caption{Overview of the controlled text generation for black-box LM with ScoPE. Starting from the neutral initial input sequence $\mathbf{x}$, ScoPE edits every $b$ tokens, $\tilde{\mathbf{y}}$, generated from $p_{LM}(\cdot | \mathbf{x})$ to $\hat{\mathbf{y}}$. $\hat{\mathbf{y}}$ has enhanced target score compared to $\tilde{\mathbf{y}}$, which means $\hat{\mathbf{y}}$ is closer to the target distribution. Edited subtext $(\mathbf{x}, \hat{\mathbf{y}})$ become a new input for the next generation step of black-box LM, guiding subsequent generations to contain the target attributes.}
     \label{fig1}
\end{figure*}

In this paper, we propose \textbf{Sco}re-based \textbf{P}rogressive \textbf{E}ditor (ScoPE) to address controlled generation for black-box language models and tackle the trade-off between control and fluency. 
ScoPE modifies the intermediate output text token blocks during the generation process of a backbone language model, ensuring that the edited text tokens contain the target attributes.
This approach effectively guides the subsequent generation to naturally include the target attributes. Since it does not access the model parameters, including the output distribution, ScoPE can be adapted to any black-box model. 
By modifying the previously generated context tokens while maintaining fluency, ScoPE ensures that the text generated afterward includes the target corpus's attributes without compromising fluency.
To effectively incorporate target attributes during the editing process, we introduce a score aligned with the attributes in the target corpus, which is computed by the score model based on a pre-trained masked language model.  
To further assist ScoPE's text modification, we additionally consider a repetition score and a task-specific score. 
Given the overall target score, ScoPE is trained with the objective of maximizing the score on text samples from a training set composed of sampling from the language models.

In our experiment for the diverse controlled generation tasks, using various corpora constructed from the Amazon Customer Reviews dataset, we comprehensively evaluate ScoPE in terms of control and fluency. We discover that ScoPE, using various black-box language models as its backbone, effectively regulates target attributes while preventing a decline in text fluency through experiments including GPT-4 evaluation. 
Additionally, our research confirms the compatibility of ScoPE with instruction prompting for control purposes. 
In tasks requiring sentiment control, ScoPE offers an effective solution to the tradeoff between fluency and control, a challenge often encountered in existing baseline models.
Furthermore, through exploring various settings of ScoPE, we affirm its proficiency in managing multi-attribute control. 

Our work presents a distinctive contribution by facilitating fluent controlled text generation utilizing black-box language models, thereby demonstrating their effectiveness and versatility for a wide range of controlled text generation tasks within the current context.

\section{Related Work}
The elements addressed by previous approaches for controlled text generation can be primarily divided into three categories: input context, weights of language models, and decoding strategy. When handling input context, the objective is to effectively incorporate target attribute information into the input of the language model (\citealp{li-liang-2021-prefix}; \citealp{Lester2021-pd_prompttun}; \citealp{Qian2022-tq_prefixcont}; \citealp{Ma2023-bnprefixcont2}). In the case of weights of language models, the approach involves fine-tuning the weights of the model, either partially or entirely, with data from the target domain (\citealp{keskarCTRL2019}; \citealp{ziegler2019finetuning}; \citealp{Lu2022-gw_quark}). Lastly, the approach to decoding strategy entails using adaptive modules, such as discriminators, that are tailored to the target domain (\citealp{Dathathri2019-tf_pplm}; \citealp{qin-etal-2020-back}; \citealp{Yang2021-rr_fudge}; \citealp{Krause2020-xi_gedi}; \citealp{liu-etal-2021-dexperts}; \citealp{Arora2022-lf_director}), or employing distributional approaches (\citealp{Deng2020-ln-residual_ebm}; \citealp{Khalifa2020-mo_dpg}; \citealp{Meng2022-wl_nado}) to perform weighted decoding of the language model. 
The practical applicability of fine-tuning approaches under black-box conditions is limited due to their dependence on model parameters. While methods of the decoding strategy can be adapted to loose black-box conditions when output distribution is approachable, they still suffer from the decreasing fluency of generated texts. 

The most prominent novelty of our work, compared to these methods, is that it secures controllability for entirely black-box LLMs, such as the ChatGPT API, where both model parameters and output logits are inaccessible.  Additionally, whereas previous works have encountered a decrease in the fluency of generated text when securing controllability, our approach addresses this issue by ensuring that the text generated after token-wise editing naturally continues from the edited previous text.

There exist studies that perform controlled generation through iterative sampling, including energy-based models, aiming to maximize scores computed from various domain-specific modules, starting from text sampled from the initial distribution (\citealp{Mireshghallah2022-vg_mixandmatch}; \citealp{Qin2022-he_cold}).
However, due to numerous iterative sampling steps, they result in significantly slower generation speeds compared to standard language model generation, reducing its practicality. About this issue, our work achieves a significant improvement in generation speed by utilizing the trained editor that only requires a few editing steps.

Recent studies introduced the method for adapting gray-box language models which only permit access to the generative distribution of the model (\citealp{ormazabal-etal-2023-comblm}; \citealp{liu-etal-2024-tuningbyproxy}). Additionally, \citet{sun2024bboxadapter} proposed a lightweight module that controls black-box language models for specific tasks. Our approach effectively adapts the black-box model from a different perspective of this concurrent work.


\section{ScoPE: Score-based Progressive Editor}
\subsection{Approach for Controlled Text Generation}
\paragraph{Overview}

In this work, we introduce ScoPE: Score-based Progressive Editor, a novel approach for controlled text generation. The primary goal is to generate a fluent continuation $\hat{\mathbf{y}}=(\hat{y}_1, ..., \hat{y}_l)$ that incorporates target attributes given an input sequence $\mathbf{x}=(x_1, ..., x_k)$. The input sequence $\mathbf{x}$ can either contain the target attributes (in-domain) or have attributes that are orthogonal or adversarial to the target (out-of-domain). Language models, when given $\mathbf{x}$ as an input, generate a continuation $\tilde{\mathbf{y}}$ that inherits the attributes of $\mathbf{x}$ due to the autoregressive nature of language modeling. This means that the generated output $(\mathbf{x}, \tilde{\mathbf{y}})$ may not always possess the desired target attributes.

To address this, we aim to perform controlled generation by editing the output $\tilde{\mathbf{y}}$ generated by the language model to incorporate the target attribute, resulting in $\hat{\mathbf{y}}$. Editing the entire $\tilde{\mathbf{y}}$ at once is challenging, so we divide the task into progressive block-wise editing during the generation process. This means that we repeatedly edit the token block generated by the language model with a short $b$ token length, and then reinsert the previously generated tokens that include the edited token block back into the input. This method allows us to steer the language model's generation step-by-step towards incorporating the target attributes effectively.

\paragraph{Block-Wise Editing}

The progressive block-wise editing approach involves dividing the continuation into blocks of $b$ tokens.
Each time $b$ generation steps are completed at time step $t$, the editor takes the generated sequence $(\mathbf{x}, \hat{\mathbf{y}}_{:t-b}, \tilde{\mathbf{y}}_{t-b:t})$ as input and modifies the current block of tokens $\tilde{\mathbf{y}}_{t-b:t} = (\tilde{y}_{t-b+1}, …, \tilde{y}_t)$ to incorporate the target domain attributes.
This iterative process ensures that the language model progressively incorporates the desired attributes in subsequent steps.

One of the key challenges in editing a token block $\tilde{\mathbf{y}}_{t-b:t}$ is the absence of gold labels that can act as $\hat{\mathbf{y}}_{t-b:t}$. In general causal language modeling, the gold label for generating a token can be considered the next token in the input $\mathbf{x}$, which is available in the training set. However, retrieving the next $b$ tokens after $\mathbf{x}$ from the training set as $\hat{\mathbf{y}}$ can be suboptimal. This is because $\tilde{\mathbf{y}}$ generated by the backbone language model often diverges from the continuation present in the training set, leading to a significant difference from $\hat{\mathbf{y}}$. In this case, $\hat{\mathbf{y}}$ is not a natural continuation from the perspective of the backbone model, leading to a decrease in the generation quality.

To overcome this, we introduce a target score that measures how well the target attribute is incorporated into the text. The objective of ScoPE is to enhance this target score while maintaining the overall structure of $\tilde{\mathbf{y}}$ for fluency. Instead of relying on labels from the training set, the target score guides the editing process, ensuring that only the essential attributes of $\tilde{\mathbf{y}}$ are transformed to align with the target domain. This approach minimizes the editor's workload and preserves the content's coherence.

The subsequent sections will delve into the training methodology for ScoPE and the formulations of the target score. Figure \ref{fig1} provides an overview of our approach.

\subsection{Training ScoPE}
\subsubsection{Preparation for ScoPE Training}
\paragraph{Training set for offline learning framework}
The inference process in ScoPE is a block-wise auto-regressive process where $\tilde{\mathbf{y}}=(\tilde{y}_1, ..., \tilde{y}_b)$, generated to continue given short sequence $\mathbf{x}$, is edited to serve as the input $\mathbf{x}$ for the next step. 
If we apply the online learning strategy to the training, $\mathbf{x}$ in the training sample $(\mathbf{x}, \tilde{\mathbf{y}})$ for the editor is constructed from the editor’s output at the previous time step. This approach, especially during the early stages of learning, cannot ensure that $\mathbf{x}$ contains the target domain attributes. Therefore, to ensure stable training, we employ the offline learning strategy. This strategy utilizes the ground truth data from the target domain as input for the next step, instead of the model’s output during training like teacher-forcing strategy (\citealp{Sutskever2014-kr_seq2seq}). In other words, to guarantee that $\mathbf{x}$ already possesses the target domain attributes, we sample $\mathbf{x}$ from the target domain data when constructing a training set for the ScoPE training.
Algorithm \ref{algorithm:trainset construction} shows an algorithm about the concrete process to construct the training set.
\paragraph{Fine-tuning pre-trained MLM}
In addition to constructing a training set for the offline learning framework, the preparation phase before training involves further tuning a pre-trained masked language model (MLM) on the target corpus. This fine-tuned MLM serves two purposes.
First, it is used as a scoring model to calculate a score that measures both the similarity to the target corpus and the level of fluency. Second, it serves as the base model for training the ScoPE model.
That is, the parameters of ScoPE are initialized with the parameters of the fine-tuned MLM. This initialization positions the initial generative distribution of ScoPE closer to the target corpus, significantly improving training stability. 

\subsubsection{Maximizing Score Disparity between Input and Edited Texts}
To ensure that the edited text achieves a higher target score than the input text, we set the objective of editor training to maximize the difference between the scores of the edited text and the input text, rather than just maximizing the score of the edited text.
To provide a more refined training signal during the learning phase, we decompose the target score of the text sequence at the token level. This token-wise target score allows the editor to receive training signals only for the token positions where edits occur, thereby enhancing training stability.
We empirically observe that the training becomes unstable when the editor receives training signals for all token positions. The training objective function $J(\theta)_t$ that should be maximized for the distribution of the editor $\theta$ at position $|\mathbf{x}|+t$ of $(\mathbf{x}, \hat{\mathbf{y}})$ is as follows:
\begin{equation}\label{eq4}
    \begin{aligned}
         J(\theta)_t &= \mathbb{E}_{\hat{y}_t\sim p_\theta(\hat{y}_t|\mathbf{x}, \tilde{\mathbf{y}})}d(\mathbf{x},\tilde{\mathbf{y}},\hat{\mathbf{y}})_t \\
         &= \sum_{\hat{y}_t\in V}p_\theta(\hat{y}_t|\mathbf{x}, \tilde{\mathbf{y}})d(\mathbf{x},\tilde{\mathbf{y}},\hat{\mathbf{y}})_t, \\
    \end{aligned}
\end{equation}
where $d(\mathbf{x},\tilde{\mathbf{y}},\hat{\mathbf{y}})_t$ is defined as follows:
\begin{equation}\label{eq5}
    \begin{aligned}
      &d(\mathbf{x},\tilde{\mathbf{y}},\hat{\mathbf{y}})_t \\
      &=\begin{cases}
 0 & \text{if } \hat{y}_t = \tilde{y}_t \\ 
 s_{t'}((\mathbf{x}, \hat{\mathbf{y}})) - s_{t'}((\mathbf{x}, \tilde{\mathbf{y}})) & \text{else},
\end{cases}
    \end{aligned}
\end{equation}
\begin{algorithm}[t]
\caption{Training set Construction}\label{algorithm:trainset construction}
\begin{algorithmic}[1]
\State \textbf{Input:} target corpus $\mathcal{X}$, backbone LM $P_{\mathrm{LM}}$,

maximum edit block size $b_{\text{max}}$, 

maximum input length $l_{\text{max}}$, 

training set size $N$
\State Define training set $\mathcal{T}=\{ \}$
\While{$|\mathcal{T}| < N$}
    \State Sample $\mathbf{x} = (x_1, \ldots, x_{l_x})$ from $\mathcal{X}$, 
    
    where $l_x \in [1, l_{\text{max}}]$
    \State Sample $\tilde{\mathbf{y}} = (y_1, \ldots, y_{l_y}) \sim P_{\mathrm{LM}}(\cdot|\mathbf{x})$, 
    
    where $l_y \in [1, b_{\text{max}}]$
    \State Construct train sample $(\mathbf{x}, \tilde{\mathbf{y}})$
    \State Append $(\mathbf{x}, \tilde{\mathbf{y}})$ to $\mathcal{T}$
\EndWhile
\State \Return $\mathcal{T}$
\end{algorithmic}
\end{algorithm}
where $s_{t'}$ denotes the decomposed score at position $t'$, $t'=|\mathbf{x}|+t$, and $V$ is a vocabulary of editor. We clip  $d(\mathbf{x},\tilde{\mathbf{y}},\hat{\mathbf{y}})_t$ within to the pre-defined range to prevent the gradient from exploding.
$J(\theta)_t$ requires computing the score for all $\hat{y}_t \in V$, which is computationally expensive and impractical. To address this issue, we approximate $J(\theta)$ by computing the expectation only for those $\hat{y}_t$ corresponding to the top-$k$ probabilities. This approximation is viable because when the editor is initialized from a fine-tuned MLM, it already possesses a reasonably sharp generative distribution from the early stages of training. Tokens with small probabilities, except for a few tokens with large probabilities, can be ignored in the calculation of expectation. In practice, we sample only one $\hat{y}_t$ from the $p_\theta(\hat{y}_t|\mathbf{x}, \tilde{\mathbf{y}})$ during training, and it shows successful results in both training and inference. When $k=1$, the gradient of $J(\theta)_t$ about $\theta$, $\nabla_\theta J(\theta)_t$ can be approximated as follows:
\begin{equation}\label{eq6}
    \begin{aligned}
         \nabla_\theta J(\theta)_t &\approx d(\mathbf{x},\tilde{\mathbf{y}},\hat{\mathbf{y}})_t \nabla_\theta p_\theta(\hat{y}_t|\mathbf{x}, \tilde{\mathbf{y}}) \\
         &= w(\mathbf{x},\tilde{\mathbf{y}},\hat{\mathbf{y}})_t \nabla_\theta \log p_\theta(\hat{y}_t|\mathbf{x}, \tilde{\mathbf{y}}) ,
    \end{aligned}
\end{equation}
where $w(\mathbf{x},\tilde{\mathbf{y}},\hat{\mathbf{y}})_t$ is defined as follows:
\begin{equation}\label{eq7}
    \begin{aligned}
         w(\mathbf{x},\tilde{\mathbf{y}},\hat{\mathbf{y}})_t = d(\mathbf{x},\tilde{\mathbf{y}},\hat{\mathbf{y}})_t p_\theta(\hat{y}_t|\mathbf{x}, \tilde{\mathbf{y}}). 
    \end{aligned}
\end{equation}
The approximated $\nabla_\theta J(\theta)_t$ is the same as the gradient of the weighted log-likelihood as follows:
\begin{equation}\label{eq8}
    \begin{gathered}
         w(\mathbf{x},\tilde{\mathbf{y}},\hat{\mathbf{y}})_t \nabla_\theta \log p_\theta(\hat{y}_t|\mathbf{x}, \tilde{\mathbf{y}}) \\
         = \nabla_\theta [w(\mathbf{x},\tilde{\mathbf{y}},\hat{\mathbf{y}})_t \log p_\theta(\hat{y}_t|\mathbf{x}, \tilde{\mathbf{y}})],
    \end{gathered}
\end{equation}
where the weighting factor $w(\mathbf{x},\tilde{\mathbf{y}},\hat{\mathbf{y}})_t$ is considered as constant. We can simply implement this by detaching the $p_\theta(\hat{y}_t|\mathbf{x}, \tilde{\mathbf{y}})$ in the weighting factor during the backward process. 

The editor can achieve more refined edits through an iterative editing process for a token block. However, when performing the $n$-th step of iteration where $n > 1$, the input for the editor, $(\mathbf{x}, \hat{\mathbf{y}}^{(n-1)})$, is generated from the editor distribution $p_\theta(\cdot|\mathbf{x}, \hat{\mathbf{y}}^{(n-2)})$, while the input for the 1st step, $\tilde{\mathbf{y}}=\hat{\mathbf{y}}^{(0)}$, is generated from the language model distribution $p_{\mathrm{LM}}(\cdot|\mathbf{x})$. As a result, these two texts can be situated in different distributions. To address this distributional mismatch, we perform N iterations of the iterative process during training. This involves sampling the edited text and using it as input to the editor again, allowing for the refinement of edits over multiple iterations. 
Finally, the loss function for a training ScoPE with training sample $(\mathbf{x}, \tilde{\mathbf{y}})=(\mathbf{x}, \hat{\mathbf{y}}^{(0)})$ can be expressed as follows:
\begin{equation}\label{eq9}
    \begin{aligned}
         &\mathcal{L}_{\mathrm{ScoPE}}\\
         &=\frac{1}{N}\sum_{i=0}^{N-1}\sum_{t=1}^{\left| \tilde{\mathbf{y}} \right|}w_t \log p_\theta(\hat{y}_t^{(i+1)}|\mathbf{x}, \hat{\mathbf{y}}^{(i)}),
    \end{aligned}
\end{equation}
where $w_t$ is detached during back-propagation and refers to as follows:
\begin{equation}\label{eq10}
    \begin{gathered}
         w_t = w(\mathbf{x},\hat{\mathbf{y}}^{(i)},\hat{\mathbf{y}}^{(i+1)})_t.
    \end{gathered}
\end{equation}
Considering the increasing training cost as $N$ grows, we fix $N=2$ in practice. 
\paragraph{Techniques for stable training}
In this work, we employ four techniques for training stability: 
1) offline learning framework, 
2) parameter initialization with MLM fine-tuned for the target corpus, 
3) providing training signals only at the positions of edited tokens, 
4) clipping the score disparity between input and edited texts.
Appendix \ref{train_alg} contains training algorithms of ScoPE.

\subsection{Score Formulation for Target Attributes}
\subsubsection{Target Score from Fine-tuned MLM}
We calculate the major target score using a masked language model (MLM) fine-tuned with target corpus. Previous research has demonstrated that an MLM trained with the objective of masked language modeling, predicting masked tokens at masked positions, can be parameterized as an implicit energy-based model (\citealp{Wang2019-je_bertmouth}; \citealp{Clark2020-rd_cloze}; \citealp{Goyal2021-rz_mh}). Also, it has been shown that fine-tuning the pre-trained MLM to the target domain improves the end-task performance in the domain (\citealp{Gururangan2020-uc_dontstop}; \citealp{Ke2023-ey_contpretraining}). By integrating the findings of these studies, we present the target score $s_{\mathrm{mlm}}(\mathbf{x})$ for a given sequence $\mathbf{x}=(x_1, ..., x_T)$ and a target domain-specific MLM $\phi$ as follows:
\begin{equation}\label{eq1}
    \begin{aligned}
         s_{\mathrm{mlm}}(\mathbf{x}) &= \sum_{t=1}^{T}s_{\mathrm{mlm}, t}(\mathbf{x}) \\
                            &= \sum_{t=1}^{T}f_\phi(x_t, h_\phi(\mathbf{x}_{\setminus t})),
    \end{aligned}
\end{equation}
where $\mathbf{x}_{\setminus t}$ is the sequence obtained by masking the position $t$ of $\mathbf{x}$, $h_\phi(\mathbf{x}_{\setminus t})$ is the representation at the position $t$ of $\mathbf{x}_{\setminus t}$ computed by $\phi$, and $f_\phi(\cdot)$ is the language modeling head function of $\phi$. In practice, the score $s_{\mathrm{mlm}, t}(\mathbf{x})$ is computed as the raw logit before entering the softmax activation function for the original token $x_t$ at the masked position, following the forward pass of the MLM. 

\subsubsection{Repetition Score}
Since the base architecture of the editor is MLM, it bidirectionally and non-autoregressively edits $\tilde{\mathbf{y}}$. However, in the non-autoregressive generation, repetitive generation remains a problem due to the multi-modality issue arising from the conditional independence assumption (\citealp{Gu2017-zi_nat}; \citealp{Zhang2022-nw_natmultimodality}). Moreover, as discussed in the existing works (\citealp{Goyal2021-rz_mh}; \citealp{Mireshghallah2022-vg_mixandmatch}), we observe that MLM may assign high scores to repetitive text, which means relying solely on the MLM score is insufficient to handle repetitive generation. To address this problem, regardless of the target attributes, we introduce a repetition score for an arbitrary sequence $\mathbf{x}=(x_1, ..., x_T)$. Repetition at the position $t$ of the sequence $\mathbf{x}$ occurs when there exists the same token as $x_t$ at different position $i$. In this case, as the distance between positions, $|i-t|$, decreases, the likelihood of an unnatural repetition increases. From this perspective, the repetition score for $\mathbf{x}$, $s_{\mathrm{rep}}(\mathbf{x})$ is described as follows:
\begin{equation}\label{eq2}
    \begin{aligned}
         s_{\mathrm{rep}}(\mathbf{x}) &= \sum_{t=1}^{T}s_{\mathrm{rep}, t}(\mathbf{x}) \\
         &= \sum_{t=1}^{T}\sum_{i=1}^{T}-\frac{\mathbbm{1}(x_i\neq x_t)}{\left| i-t \right|},
    \end{aligned}
\end{equation}
where $\mathbbm{1}(\cdot)$ denotes the indicator function.

\subsubsection{Task-specific Score}
In addition to the MLM score, other task-specific scores such as raw logits from a discriminator for the target attribute, automatic evaluation metric, or human feedback can also be combined with the target score.
Motivated by the previous work (\citealp{Mireshghallah2022-vg_mixandmatch}), the target score is augmented through a linear combination of domain-specific scores. For example, the integration of $s_{\mathrm{mlm}}$, $s_{\mathrm{rep}}$, and $s_{\mathrm{disc}}$, is computed as follows:
\begin{equation}\label{eq3}
    \begin{aligned}
         s_{\mathrm{total}, t} = s_{\mathrm{mlm}, t} + \alpha \cdot s_{\mathrm{rep}, t} + \beta \cdot s_{\mathrm{disc}},
    \end{aligned}
\end{equation}
where $\alpha$ and $\beta$ are the scaling factors for scores, and $s_{\mathrm{total}, t}$ is the overall score at position $t$. As $s_{\mathrm{disc}}$ is a sequence-wise score, we assign the same score $s_{\mathrm{disc}}$ for all positions.

\section{Experimental Setup}
We conduct a controlled generation evaluation using the Amazon Customer Reviews dataset\footnote{https://s3.amazonaws.com/amazon-reviews-pds/readme.html}.
The dataset includes multiple attributes related to the review style, the categories corresponding to the type of reviewed product, and the sentiments aligned with the product rating. Consequently, this dataset is well-suited for the creation of diverse corpora that contain specific attributes. We classify the dataset corpus based on category and sentiment. For categories, we construct four distinct corpora: Camera, Videogame, Grocery, and Music. In terms of sentiment, samples rated with 5 stars are compiled into a positive corpus, while those rated with 1 star formed a negative corpus. Details on the dataset statistics are discussed in Appendix \ref{data_statistics}.

Using these corpora, we evaluate controlled generation tasks for both category and sentiment attributes. In each task, we test the controllability and fluency of generated continuations when fixing one target attribute among the attributes relevant to the task, starting from an input text with arbitrary attributes. 
Experimental details for our experiments are mentioned in Appendix \ref{experi_details}.
\begin{figure*}[t]
     \centering
     \includegraphics[width=\textwidth]{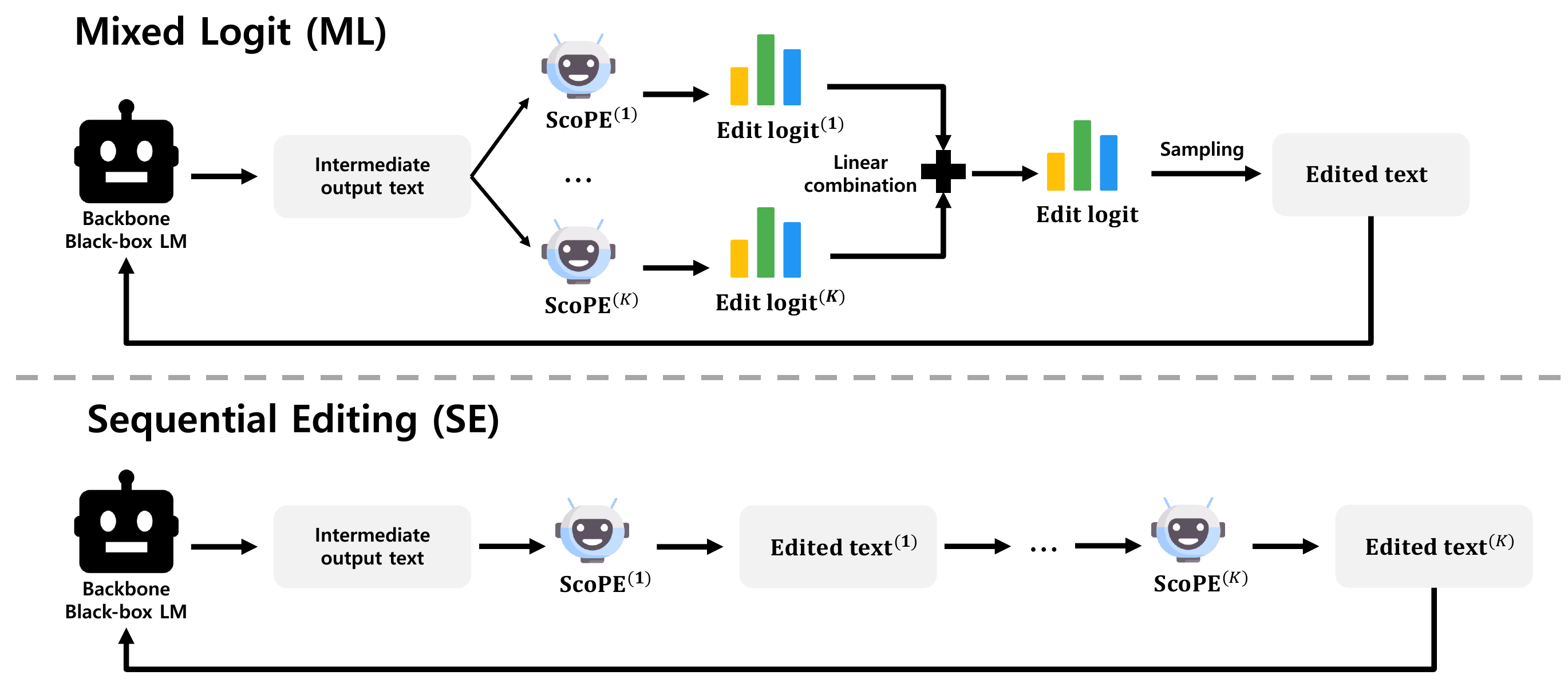}
     \caption{Overview of the mechanism of the two approaches for the multi-attributes controlled generation using ScoPE: mixed logit (ML), sequential editing (SE).}
     \label{fig_ml_se}
\end{figure*}
\begin{table*}[t]
\centering
{\footnotesize\resizebox{\textwidth}{!}{
\renewcommand{\arraystretch}{1.3}
\begin{tabular}{l||cccc|cccc}
\toprule
 \multirow{2}{*}{\textbf{Methods}} & \multicolumn{4}{c|}{$\textbf{PPL}$ $\downarrow$} & \multicolumn{4}{c}{\textbf{MAUVE} $\uparrow$}          \\
             &  \textbf{Camera} & \textbf{Videogame} & \textbf{Grocery} & \textbf{Music} & \textbf{Camera} &\textbf{Videogame} & \textbf{Grocery} & \textbf{Music}   \\ \hline
\texttt{davinci-002} &  25.47& 27.19& 23.67& 28.24
& 0.7840& 0.2200&  0.1098& 0.0466
\\
 ScoPE ($N$=5) & \textbf{ 14.93}&  \textbf{16.71}& \textbf{16.98}& \textbf{19.88}& \textbf{0.8850}& \textbf{0.8848}&  \textbf{0.8418}& \textbf{0.7432}\\ \hline
\texttt{babbage-002} &  34.57&  37.14& 30.96& 38.57
& 0.7312& 0.2386&  0.1122& 0.0516
\\
 ScoPE ($N$=5) &  \textbf{15.21}&  \textbf{16.78}&\textbf{17.17}& \textbf{19.95}&\textbf{0.8888}& \textbf{0.8658}&  \textbf{0.8431}& \textbf{0.7620}\\
\bottomrule
\end{tabular}
}}
\caption{Experimental results for category controlled generation targeting the Camera attribute. \texttt{davinci-002} and \texttt{babbage-002} are utilized for backbone model at the generation process. $N$ denotes the number of iterative edits performed on the input text.}
\label{table2}
\end{table*}
\begin{table*}[t]
\centering
{\footnotesize\resizebox{\textwidth}{!}{
\renewcommand{\arraystretch}{1.3}
\begin{tabular}{l||cccc|cccc}
\toprule
 \multirow{2}{*}{\textbf{Results}} & \multicolumn{4}{c|}{\textbf{Fluency}} & \multicolumn{4}{c}{\textbf{Controllability}}          \\
             &  \textbf{Camera} & \textbf{Videogame} & \textbf{Grocery} & \textbf{Music} & \textbf{Camera} &\textbf{Videogame} & \textbf{Grocery} & \textbf{Music}   \\ \hline
Win  &  \textbf{0.82}& \textbf{0.72}& \textbf{0.61}& \textbf{0.70}& \textbf{0.85}& \textbf{0.90}&  \textbf{0.88}& \textbf{0.83}\\
Lose &  0.18&  0.26& 0.34& 0.27& 0.12& 0.00&  0.00& 0.00\\
Draw &  0.00&  0.02& 0.05& 0.03& 0.03& 0.10&  0.12& 0.17\\ 
\bottomrule
\end{tabular}
}}
\caption{Comparative evaluation for the fluency and controllability of Camera attribute between text generated by ScoPE using \texttt{Davinci-002} as the backbone model and text generated by Davinci-002 alone (assessed by GPT-4). Win / Lose / Draw indicates the
percentage of times our method wins, loses or draws against \texttt{Davinci-002} respectively.}
\label{table_gpt4}
\end{table*}
\begin{table*}[t]
\centering
{\footnotesize\resizebox{\textwidth}{!}{
\renewcommand{\arraystretch}{1.3}
\begin{tabular}{l||cccc|cccc}
\toprule
 \multirow{2}{*}{\textbf{Methods}} & \multicolumn{4}{c|}{$\textbf{PPL}$ $\downarrow$} & \multicolumn{4}{c}{\textbf{MAUVE} $\uparrow$}          \\
             &  \textbf{Camera} & \textbf{Videogame} & \textbf{Grocery} & \textbf{Music} & \textbf{Camera} &\textbf{Videogame} & \textbf{Grocery} & \textbf{Music}   \\ \hline
\texttt{gpt-3.5-turbo-0613} & \textbf{5.15}& \textbf{5.05}& \textbf{4.97}& \textbf{5.16}
& 0.2342& 0.0508&  0.0269& 0.0144
\\
 $+$ Instruction &  6.15&  6.4& 6.67& 6.65
& 0.3109& 0.2211&  0.2151& 0.1653
\\
ScoPE ($N$=5) &  7.09&  8.28& 8.25& 9.65
& 0.3446& 0.2195&  0.2285& 0.0753
\\
$+$ Instruction &  6.63&  6.9& 6.89& 7.15& \textbf{0.4302}& \textbf{0.3618}&  \textbf{0.3886}& \textbf{0.3218}\\ 
\bottomrule
\end{tabular}
}}
\caption{Experimental results for category controlled generation utilizing the \texttt{gpt-3.5-turbo-0613} (ChatGPT). The target attribute is Camera.}
\label{table3}
\end{table*}
\begin{table*}[t]
\centering
{\resizebox{\textwidth}{!}{
\renewcommand{\arraystretch}{1.2}
\begin{tabular}{l||cc|cc|cc||cc|cc|cc}
\toprule
\multirow{3}{*}{\textbf{Methods}} &\multicolumn{6}{c||}{\textbf{Positive targeting}} & \multicolumn{6}{c}{\textbf{Negative targeting}} \\ \cline{2-13}
 & \multicolumn{2}{c|}{$\textbf{PPL}$ $\downarrow$} & \multicolumn{2}{c|}{\textbf{Acc. 1} $\uparrow$} & \multicolumn{2}{c||}{\textbf{Acc. 2} $\uparrow$} & \multicolumn{2}{c|}{$\textbf{PPL}$ $\downarrow$} & \multicolumn{2}{c|}{\textbf{Acc. 1} $\uparrow$} & \multicolumn{2}{c}{\textbf{Acc. 2} $\uparrow$} \\
& \textbf{Pos.} & \textbf{Neg.} & \textbf{Pos.} & \textbf{Neg.} & \textbf{Pos.} & \textbf{Neg.} & \textbf{Pos.} & \textbf{Neg.} & \textbf{Pos.} & \textbf{Neg.} & \textbf{Pos.} & \textbf{Neg.} \\ \hline
DExperts & 192.07& 123.82& \textbf{99.48}& 81.8& 80.26& 61.82 & 44.83& 69.03& \textbf{91.38}& \textbf{99.37}& \textbf{90.52}& \textbf{97.87}\\
GeDi & 90.59& 74.19& 97.65& 67.52& 67.20& 45.64& 80.51& 94.64& 73.82& 95.58& 74.91& 91.49\\
Mix\&Match & 15.94& 16.71& 98.69& 81.41& 55.20& 17.66& 16.4& 16.55& 55.82& 92.57& 20.14& 66.65\\ \hline
ScoPE ($N$=10)& 12.01& 12.84& 95.57& 64.29& 77.04& 41.04& \textbf{12.6}& \textbf{12.91}& 54.41& 88.01& 55.9& 87.88\\
 $+$ EDG & \textbf{11.99}& \textbf{12.55}& 99.00& \textbf{84.55}& \textbf{86.19}& \textbf{63.63} & 12.79& 13.08& 79.59& 95.4& 81.84& 94.43\\
\bottomrule
\end{tabular}
}}
\caption{Experimental results for sentiment controlled text generation comparing with prior works. GPT2-XL is utilized for both ScoPE trainset construction and the backbone model for the ScoPE generation process. The accuracy from the discriminator used to ScoPE training is denoted as \textbf{Acc. 1}, while the accuracy from the discriminator not used to ScoPE training is denoted as \textbf{Acc. 2}.
}
\label{table5}
\end{table*}
\begin{table*}[t]
\centering
{\footnotesize\resizebox{\textwidth}{!}{
\renewcommand{\arraystretch}{1.3}
\begin{tabular}{l||cccc|cc|cc}
\toprule
 \multirow{2}{*}{\textbf{Methods}} & \multicolumn{4}{c|}{$\textbf{MAUVE}$ $\uparrow$} & \multicolumn{2}{c|}{\textbf{Acc. 1} $\uparrow$} & \multicolumn{2}{c}{\textbf{Acc. 2} $\uparrow$}         \\
             &  \textbf{Camera} & \textbf{Videogame} & \textbf{Grocery} & \textbf{Music} & \textbf{Pos.} &\textbf{Neg.} & \textbf{Pos.} & \textbf{Neg.}   \\ \hline
\texttt{davinci-002} &  0.7840& 0.2200& 0.1098& 0.0466& 86.99&23.01&  51.62& 8.09\\ \hline
 ScoPE - MS ($N$=5) & 0.8553&  \textbf{0.8431}& 0.7632& \textbf{0.7707}& \textbf{99.63}& \textbf{84.87}& \textbf{88.79}& \textbf{74.09}\\
ScoPE - ML ($N$=5) &  \textbf{0.8846}&  0.8381& \textbf{0.8288}& 0.6972& 92.03& 44.04& 72.78& 28.17\\
 ScoPE - SE ($N$=5) &  0.8474&  0.8116&0.7381& 0.6520&96.74& 65.17& 82.18& 48.03\\
\bottomrule
\end{tabular}
}}
\caption{Experimental results for multi-attribute controlled generation targeting both the Positive and Camera attributes. \texttt{davinci-002} is utilized for the backbone model at the generation process.}
\label{table_multi_camera_pos}
\end{table*}
\subsection{Category Controlled Generation}
In the category controlled generation task, we employ Perplexity (PPL) calculated from LLaMA2-13B as the metric for measuring fluency. For assessing controllability, we employ the MAUVE (\citealp{Pillutla2021-bc_mauve}) metric. MAUVE calculates the distance between the approximated distributions of generated texts and reference texts. A detailed description of the empirical study examining the suitability of MAUVE as a metric for controllability is presented in the Appendix \ref{mauve}. 
Recently, it has become increasingly recognized that using GPT-4 to assess the quality of text correlates more closely with human preferences compared to existing evaluation metrics (\citealp{liu-etal-2023-geval}; \citealp{controlled_arithmetic}).
Consequently, in addition to automatic metrics, 
we conduct a comparative evaluation between ScoPE and the backbone LM using GPT-4 to provide a more comprehensive assessment that aligns with human preferences. The prompts injected into GPT-4 for the evaluation can be found in the Appendix \ref{gpt4evalprompt}.

In this task, control over the target attributes was executed under the corresponding target corpus, not under the specific keywords.
Consequently, traditional keyword-focused topic control baselines are not suitable for comparison in this task. Hence, our evaluation in the category controlled generation task explores adaptability with various black-box models rather than comparison with existing baselines. 
We compose a training set from the GPT2-XL and apply it to controlled generation using two black-box API backbones: \texttt{davinci-002} and \texttt{babbage-002}. Moreover, we further refine ScoPE, which was initially trained with a GPT2-XL composed training set, by tuning it on a smaller training set derived from the black-box instruct LLM API: \texttt{gpt-3.5-turbo-0613}. We then evaluate various settings, including scenarios where ScoPE is combined with the instruction prompting of the specified backbone model.

\subsection{Sentiment Controlled Generation}
In the sentiment controlled generation task, we use Perplexity (PPL) as the metric for measuring fluency and employ two pre-trained sentiment classifiers as metrics for assessing controllability. For evaluating controllability, the accuracy concerning the target sentiment of each classifier is utilized. Among these classifiers, one is used for score calculation during the training of ScoPE (\citealp{hartmann2023_sentclassifier1}). We utilize the raw logits of the classifier as the additional target score $s_{\mathrm{disc}}$, which is denoted as EDG (External Discriminator Guidance). Since this might result in an overfitted outcome for this classifier, we adopt an additional pre-trained sentiment classifier of a different type to check overfitting (\citealp{hartmann2021_sentclassifier2}). 

We conduct the comparison of ScoPE with existing baselines that manipulate the output distribution of the backbone language model: GeDi, DExperts, (\citealp{Krause2020-xi_gedi}; \citealp{liu-etal-2021-dexperts}) or sample the generated text through iterative steps: Mix\&Match (\citealp{Mireshghallah2022-vg_mixandmatch}). To ensure fair performance comparison in terms of the backbone model, we use GPT2-XL both for the training set composition and the backbone language model in the generation process. 

\subsection{Multi-Attribute Controlled Generation}
Our proposed method, ScoPE, can be designed to control multiple attributes simultaneously. In this work, we explore a multi-attribute control methodology with below three variants of ScoPE.
\begin{itemize}[leftmargin=*, nolistsep]
    \item{ScoPE-MS (Mixed Score)}: During the training process of ScoPE, scores for various attributes are combined in the form of Eq. \ref{eq3} to construct an overall target score.
    \item{ScoPE-ML (Mixed Logit)}: During the generation process, when calculating logits for editing, the logits from editors targeting each attribute are combined through linear combination, resulting in a new calculated logit. 
    \item{ScoPE-SE (Sequential Editing)}: During the generation process, various editors sequentially perform token-level edits in the editing phase.
\end{itemize}
Fig. \ref{fig_ml_se} illustrates the overall editing mechanisms of ScoPE for multi-attribute controlled generation using both the ML and SE approaches for handling K different attributes. We evaluate ScoPE's ability to control specific categories and sentiments simultaneously. Detailed descriptions of ML and SE can be found in the Appendix \ref{ml_se_descrip}.

\section{Results}
\subsection{Category Controlled Generation}
Table \ref{table2} presents the evaluation results of controlled generation for two GPT3 base model APIs, targeting the Camera attribute.
The improvements in both PPL and MAUVE, regardless of the type of backbone model, demonstrate ScoPE's adaptability to various black-box models. Moreover, the ability to enhance the fluency of backbone models while ensuring controllability over target attributes indicates that ScoPE's controllability is not achieved at the expense of fluency. This highlights the balanced effectiveness of ScoPE in managing both aspects simultaneously.

Table \ref{table_gpt4} presents a comparative experiment evaluation for fluency and controllability. Regardless of the input text's attributes, texts generated via ScoPE were evaluated as more fluent than those produced solely by the backbone LLM ("Win" > 50). This suggests that despite utilizing an editor with a significantly smaller parameter size compared to the backbone model, ScoPE's generation is indeed more fluent. This trend aligns with the results of PPL. In terms of controllability, ScoPE also received better evaluations in all settings compared to the backbone model. In particular, ScoPE generation was evaluated as superior even when the input attribute was Camera, so that the continuation generated by the backbone model was likely to belong to the camera category.

Table \ref{table3} presents the evaluation results of controlled generation for the ChatGPT API, targeting the Camera attribute. Focusing on the fact that ChatGPT is an instruct LLM, we confirm that the ScoPE generation method can be effectively combined with instruction prompting. It demonstrates significantly higher performance in terms of control while preserving the fluency of a large-scale model, as compared to when each method is used independently. This underscores the compatibility between the commonly used prompting methods in LLMs and ScoPE. The instruction prompts for ChatGPT can be found in Appendix \ref{chatgptinst}.

\subsection{Sentiment Controlled Generation}
Table \ref{table5} presents the evaluation results of comparisons with prior works in sentiment controlled generation targeting the positive attributes. Although control over sentiment is not always predominant, ScoPE shows significantly improved results in terms of fluency compared to DExperts and GeDi. This can be seen as a meaningful resolution to the trade-off that typically arises in existing methodologies manipulating the output distribution of the backbone models, where increasing controllability often leads to a reduction in fluency. 

When compared with Mix\&Match, this approach also preserves a degree of fluency, as it employs token-level iterative sampling instead of directly altering the distribution. However, it is observed that while control performance is ensured for the used sentiment classifier, it declines for other classifiers, suggesting that overfitting has occurred specifically for the chosen sentiment classifier. The improved accuracy of the external discriminator not used in ScoPE training shows that the ScoPE is not overfitted to the discriminator used in training. The results of targeting the negative attributes can be found in Appendix \ref{additional_results}.

\subsection{Multi-Attribute Controlled Generation}
Table \ref{table_multi_camera_pos} presents results from utilizing ScoPE in multi-attribute control settings, targeting both Camera and Positive attributes simultaneously. The outcomes suggest that all three settings successfully target both attributes. In the case of MS, it particularly demonstrates superior control over the sentiment attribute compared to the other two settings. While ML and SE show slightly lower performance than MS, they hold an advantage in that they can freely use pre-tuned editors for single attribute targeting in a plug-and-play manner. Notably, ML offers the highest degree of freedom by allowing the adjustment of each attribute's proportion.

Additionally, the control explorations using LLaMA2-7B model as the black-box backbone model and the the diversity evaluation of the texts are presented in Appendix \ref{additional_results}. Including ablation studies, the analysis of the inference cost and fluency are presented in Appendix \ref{analysis}. Appendix \ref{Generated Samples} shows the generated samples of ScoPE.

\section{Conclusion}
We present ScoPE which guides the generation process of a backbone language model to improve target domain scores, enabling fluent controlled text generation. ScoPE effectively addresses the challenges associated with the black-box scenario and the trade-off between controllability and fluency. Furthermore, it demonstrates its pragmatic utility when incorporated within various large language model APIs, manifesting its tangible applicability in real-world contexts. We believe ScoPE, combined with improved functions, can be effectively utilized for the adaptation of large language models to specific domains or tasks in future works..

\section*{Limitations}
In this section, we mention several limitations of our method. During the training process, there is a memory and time cost incurred due to the need for masking at each position within the sequence when calculating the MLM score. Additionally, to understand the target domain attributes and analyze the distribution of text generated by the backbone language model, a certain amount of target domain data and samples from the backbone language model are required. In a few-shot setting, additional methods would be necessary for future work. If the target domain and the domain to which the input belongs are too different, the burden on the editor during editor training becomes significant. As a result, the frequency of modifications by the editor increases significantly, leading to instability in the training process and a substantial increase in cost. In conclusion, considering the challenging nature of training and inference, addressing the issue of editor instability should be a key topic for future work. Furthermore, ScoPE uses only the replacement option among the possible edition options (replacement, insertion, deletion). Architecture or training methods for performing various edition options could be improvement directions of future works. We discuss additional limitations about the inference cost and fluency of the generated text in Appendix \ref{analysis}.

\section*{Ethics Statement}
In this section, we aim to address the ethical issues we perceive in our work. The most significant concern pertains to the potential of our controlled text generation task to target morally objectionable attributes. The ability of ScoPE to generate text limited to a desired target domain, regardless of the input's originating domain, is an important issue related to the \textbf{misalignment of large language models}. Referred to as "Jailbreak," this topic deals with the phenomenon where language models become misaligned by specific contextual inputs, resulting in the generation of non-preferred and potentially harmful text. Our research involves guiding the context towards the target domain during the generation process of the language model, which could be exploited to induce misalignment in the language model by designating the target domain as the domain of misalignment. To mitigate such misuse, we propose leveraging misalignment as a shield against misalignment attacks by establishing ScoPE's target as the generation of only preferred text, thereby preventing the occurrence of misalignment.

\section*{Acknowledgements}
This work was supported by the National Research Foundation of Korea (NRF) grants funded by the Korea government (Ministry of Science and ICT, MSIT) (2022R1A3B1077720), Institute of Information \& Communications Technology Planning \& Evaluation (IITP) grants funded by the Korea government (MSIT) (RS-2021-II211343: Artificial Intelligence
Graduate School Program (Seoul National University) and 2022-0-00959), Samsung Electronics Co., Ltd (IO221213-04119-01), and the BK21 FOUR
program of the Education and Research Program for Future ICT Pioneers, Seoul National University in 2024.

\bibliography{scope}

\clearpage
\appendix

\section{Training Algorithms for ScoPE} \label{train_alg}

\begin{algorithm}
\caption{MLM Fine-Tuning}
\begin{algorithmic}[1]
\State \textbf{Input:} pre-trained MLM $\psi$, 

fine-tuned MLM $\phi$, target corpus $\mathcal{X}$, 

MLM loss function $\mathcal{L}_{\mathrm{MLM}}$
\State Initialize $\phi$ with $\psi$
\For {$\mathbf{x}$ in $\mathcal{X}$}
    \State Update $\phi$ by $\nabla_\phi\mathcal{L}_{\mathrm{MLM}}(\mathbf{x})$
\EndFor
\State \Return $\phi$
\end{algorithmic}
\end{algorithm}

\begin{algorithm}
\caption{Training ScoPE} 
\begin{algorithmic}[1]
\State \textbf{Input:} training set $\mathcal{T}$, fine-tuned MLM $\phi$, 

ScoPE $\theta$, clip scale $c$
\State Initialize $\theta$ with $\phi$
\For {$(\mathbf{x}^{(i)}, \tilde{\mathbf{y}}^{(i)})$ in $T$}
    \State Sample $\hat{\mathbf{y}}^{(i)} \sim p_\theta(\cdot|\mathbf{x}^{(i)}, \tilde{\mathbf{y}}^{(i)})$
    \State Compute $d(\textbf{x},\tilde{\textbf{y}},\hat{\textbf{y}})_t$ for $0 \leq t < |\tilde{\mathbf{y}}^{(i)}|$
    \State Clip $d(\textbf{x},\tilde{\textbf{y}},\hat{\textbf{y}})_t$ within the range $(-c, c)$
    \State Compute $\mathcal{L}_{\mathrm{ScoPE}}$
    \State Update $\theta$ by $\nabla_\theta\mathcal{L}_{\mathrm{ScoPE}}$
\EndFor
\State \textbf{return} $\theta$
\end{algorithmic}
\end{algorithm}

\section{Data Statistics} \label{data_statistics}
Table \ref{table_datastatistics} shows the data statistics for all domains used in our work. When training and evaluating ScoPE combined with \texttt{gpt-3.5-turbo-0613} API, we utilize about 5\% of the data corpus of the Camera corpus.

\section{Experimental Details} \label{experi_details}
The test set created for evaluation consists of input texts of 32 tokens and generated continuation texts of 128 tokens. During test set creation, the block size b for editing was fixed at 16. For MAUVE evaluation, we compute the performance of MAUVE using five seeds for k-means clustering and evaluate it as the average. A reference set for MAUVE evaluation is constructed from the target corpus.

Table \ref{table_hyperparam_mlm} presents the hyperparameters for fine-tuning the pre-trained MLM. Table \ref{table_hyperparam_scope} presents the hyperparameters for training ScoPE. All training and fine-tunings are conducted with four GPU on our machine (GPU: NVIDIA V100). Irrespective of the domain, during the first stage of ScoPE training, the additional fine-tuning of the pre-trained MLM has been observed to transpire at a rate of approximately 1.8 updates per second. The subsequent stage, involving ScoPE loss training, proceeds at an approximate rate of 0.45 updates per second. Given that the camera domain, characterized by the largest target dataset size, requires the most extensive training, the complete training process for this domain takes approximately 29h 52m time. Our code is based on \texttt{FairSeq} (\citealp{Ott2019-pr_fairseq}). The pre-trained MLM used for acquiring the base models and score models of ScoPE in all tasks is RoBERTa-\texttt{base} (\citealp{Liu2019-oc_roberta}).We fix the output text decoding strategy of the ScoPE editor to ancestral sampling in all evaluations.

\section{Empirical Study for MAUVE as Controllability Metric} \label{mauve}
\begin{table}[t]
\centering
{\normalsize{
\renewcommand{\arraystretch}{1.2}
\begin{tabular}{l|cc}
\toprule
\textbf{Domains} & \textbf{\# Tokens} & \textbf{\# Samples} \\ \hline
Camera & 178M & 1.80M \\
Videogame & 222M & 1.79M \\
Grocery & 141M & 2.40M \\
Music & 165M & 1.80M \\
Positive & 64.4M & 647K \\
Negative & 60.9M & 613K \\
\bottomrule
\end{tabular}
}}
\caption{Data statistics for each domain data corpus.}
\label{table_datastatistics}
\end{table}
\begin{table}[t]
\centering
{\small \resizebox{\columnwidth}{!}{
\renewcommand{\arraystretch}{1.3}
\begin{tabular}{l|cccc}
\toprule
\multirow{2}{*}{\textbf{Source}} & \multicolumn{4}{c}{\textbf{Reference}} \\
                  &  Camera    &  Videogame     & Grocery     &   Music  \\ \hline
     Camera       &  \textbf{0.9441}   &  0.1912   &  0.0653	& 0.0322    \\
     Videogame    &  0.2582	&\textbf{0.9056}	&0.0725	&0.0931    \\
      Grocery            &   0.0549	& 0.0524  &	\textbf{0.9542} & 0.0322    \\
       Music           &  0.0327 &	0.1008	& 0.0378	& \textbf{0.9563} \\
\bottomrule
\end{tabular}
}}
\caption{Mauve of categorical domains. "Source" refers to the source domain, and "Reference" refers to the reference domain.}
\label{table_mlmcates_mauve}
\end{table}
In this section, we present our empirical studies on MAUVE, which are evaluation metrics used for scoring target attributes. 
Referring to the previous work which already demonstrated that MAUVE effectively indicates sentimental features in the text (\citealp{Pimentel2022-cu_mauveanal}), we investigate whether it also effectively indicates categorical features. 
Since MAUVE considers the comprehensive attributes of the target corpus, it evaluates not only the category attribute but also other attributes like review style. To tackle this issue, we split the test set for each domain into two parts, designating one as the source set for MAUVE calculation and the other as the reference set. We measure the scores for all combinations of domains. 

Table \ref{table_mlmcates_mauve} demonstrates that MAUVE effectively indicates categorical attributes by showing significantly higher values only when the source and reference domains are the same. Notably, even though sets with different category attributes share commonalities in other attributes, the differences in MAUVE scores are pronounced. This suggests that the category attribute, compared to style or other attributes, possesses more distinct characteristics, allowing for effective differentiation through MAUVE measurements.

In addition to MAUVE, we also used GPT-4 as a means to verify the controllability of ScoPE by comparing which of the ScoPE-generated texts or the backbone model (\texttt{Davinci-002}) generated texts are closer to the target category for the same input text in terms of continuations. The results of this evaluation align with the MAUVE results, particularly highlighting that the control over the Music category is relatively lacking, which is corroborated by the GPT-4 evaluation results. This adds further justification for using the MAUVE metric as a measure of controllability in our work.

\section{Prompts for Comparative Evaluation using GPT-4} \label{gpt4evalprompt}
Table \ref{table_gpt4eval_prompt_fluency} shows the content of the prompts injected to GPT-4 for the comparative evaluation of fluency. Table \ref{table_gpt4eval_prompt_cont} shows the content of the prompts injected to GPT-4 for the comparative evaluation of controllability.

\section{Prompts for Controllability Instruction of ChatGPT} \label{chatgptinst}
Table \ref{table_chatgpt_prompt_cont} shows the content of the prompts injected to ChatGPT for the instruction of controllability to Camera domain.
\begin{table}[t]
\centering
{\footnotesize\resizebox{\columnwidth}{!}{
\renewcommand{\arraystretch}{1.2}
\begin{tabular}{l||cc|cc|cc}
\toprule
\multirow{2}{*}{\textbf{Methods}} & \multicolumn{2}{c|}{$\textbf{PPL}$ $\downarrow$} & \multicolumn{2}{c|}{\textbf{Acc. 1} $\uparrow$} & \multicolumn{2}{c}{\textbf{Acc. 2} $\uparrow$} \\
& \textbf{Pos.} & \textbf{Neg.} & \textbf{Pos.} & \textbf{Neg.} & \textbf{Pos.} & \textbf{Neg.} \\ \hline
LLaMA2-7B & \textbf{11.49}& 12.73
& 88.75& 26.56
& 53.74& 8.22
\\
ScoPE ($N$=1) & 13.53& 14.45
& 93.86& 44.64 & 70.67& 24.47
\\
 $+$ EDG & 13.52& 14.81
& 97.86& 71.76
& 80.12& 50.52
\\
ScoPE ($N$=5) & 11.85& 12.63
& 95.35& 59.23
& 76.76& 38.02
\\
 $+$ EDG & 11.81& 12.40& 98.97& 82.74
& 87.19& 62.38
\\
ScoPE ($N$=10) & 11.76& 12.67
& 95.64& 60.70& 77.61& 40.06
\\
$+$ EDG & 11.64& \textbf{12.39}& \textbf{99.10}& \textbf{83.76}& \textbf{87.22}& \textbf{65.29}\\
\bottomrule
\end{tabular}
}}
\caption{Experimental results for sentiment controlled text generation targeting the positive attribute. LLaMA2-7B is utilized for both trainset construction and the backbone model for the generation process. The accuracy from the discriminator used to ScoPE training is denoted as \textbf{Acc. 1}, while the accuracy from the discriminator not used to ScoPE training is denoted as \textbf{Acc. 2}. EDG refers to external discriminator guidance.}
\label{sentiment_llama2_pos}
\end{table}

\section{Detailed Descriptions of ScoPE-ML and ScoPE-SE for Multi-attribute Controlled Generation} \label{ml_se_descrip}
In the ML method, the final edit logits are calculated through a linear combination of the logits from editors tuned to target different attributes. This process allows for the adjustment of the relative importance of each attribute by manipulating the weights in the linear combination. For instance, in the Camera-Positive multi-attributes control scenario, we set the weight for the Camera attribute at 10.0 and for the positive attribute at 1.0.

The SE method involves using editors, each tuned for different attributes, in a sequential manner for editing. Consequently, the overall editing process consists of K editing sub-processes. In this approach, the order in which editors are applied significantly impacts the overall editing performance. Through empirical studies, we have found that placing broader attributes like sentiment earlier in the sequence enhances performance. In practice, for the Camera-Positive control task, we first used the editor for the positive attribute, followed by the editor for the Camera attribute.

\section{Additional Results} \label{additional_results}
\begin{table}[t]
\centering
{\footnotesize\resizebox{\columnwidth}{!}{
\renewcommand{\arraystretch}{1.2}
\begin{tabular}{l||cc|cc|cc}
\toprule
\multirow{2}{*}{\textbf{Methods}} & \multicolumn{2}{c|}{$\textbf{PPL}$ $\downarrow$} & \multicolumn{2}{c|}{\textbf{Acc. 1} $\uparrow$} & \multicolumn{2}{c}{\textbf{Acc. 2} $\uparrow$} \\
& \textbf{Pos.} & \textbf{Neg.} & \textbf{Pos.} & \textbf{Neg.} & \textbf{Pos.} & \textbf{Neg.} \\ \hline
LLaMA2-7B & \textbf{11.5}& 12.71
& 11.26& 73.83
& 12.79& 68.81
\\
ScoPE ($N$=1) & 13.85& 14.34
& 30.87& 83.26
& 36.89& 82.21
\\
 $+$ EDG & 13.83& 14.07
& 65.47& 91.56
& 66.72& 89.67
\\
ScoPE ($N$=5) & 12.4& 12.69
& 49.52& 87.4
& 53.59& 87.59
\\
 $+$ EDG & 12.22& 12.58
& 78.19& 95.75
& 79.36& 94.47
\\
ScoPE ($N$=10) & 12.4& 12.59
& 51.84& 89.03
& 55.71& 88.93
\\
$+$ EDG & 12.26& \textbf{12.35}& \textbf{80.33}& \textbf{96.05}& \textbf{80.93}& \textbf{94.78}\\
\bottomrule
\end{tabular}
}}
\caption{Experimental results for sentiment controlled text generation targeting the negative attribute. The accuracy from the discriminator used to ScoPE training is denoted as \textbf{Acc. 1}, while the accuracy from the discriminator not used to ScoPE training is denoted as \textbf{Acc. 2}. EDG refers to external discriminator guidance.}
\label{sentiment_llama2_neg}
\end{table}

\begin{table}[t]
\centering
{\normalsize{
\renewcommand{\arraystretch}{1.2}
\begin{tabular}{l|ccc}
\toprule
\textbf{Methods} & \textbf{Dist-1} & \textbf{Dist-2} & \textbf{Dist-3}  \\ \hline
\texttt{Davinci-002} & 0.675 & 0.927 & 0.973 \\
ScoPE ($N$=5) & 0.649 & 0.919 & 0.971 \\
\bottomrule
\end{tabular}
}}
\caption{Results of distinctness measurement for texts generated by ScoPE with \texttt
{Davinci-002} as the backbone model and texts generated solely by \texttt{Davinci-002.}}
\label{table_diversity}
\end{table}
\paragraph{Results utilizing LLaMA2-7B} 
To explore the most basic setting where the backbone language models used for training set composition and in the generation process are the same, we use the relatively large-scale model LLaMA2-7B to construct the training set for ScoPE, then apply this ScoPE to controlled generation using the same type of backbone language model. Table \ref{category_llama2_others} presents the evaluation results of controlled generation targeting the Camera attribute. As the number of iterations $N$ increases, the controllability of ScoPE (MAUVE) usually improves along with the fluency (PPL). These results indicate that the ScoPE framework is capable of generating texts that are not only fluent but also precisely controlled. Table \ref{sentiment_llama2_pos} and \ref{sentiment_llama2_neg} present the experimental results for sentiment controlled text generation tasks targeting each sentiment domain. To examine the impact of using an external discriminator, we denote the cases where only MLM score and repetition score are used, as in the baseline, as ScoPE, and the cases where the external discriminator is included in score computation as EDG (External Discriminator Guidance). In all cases using ScoPE, we observe improved scores and MAUVE, demonstrating the successful steering of the language model's controlled generation in the target attributes. Moreover, the improved accuracy of the external discriminator not used in ScoPE training shows that the ScoPE is not overfitted to the discriminator used in training.

\paragraph{Diversity evaluation}
We assess the diversity of the texts generated by ScoPE using Davinci-002 as the backbone model and those directly generated by Davinci-002, employing the distinctness metric. Distinctness was calculated by measuring the unique number of unigrams, bigrams, and trigrams in each generated sample and dividing this by the total number of (uni- or bi- or tri-)grams in the sample (\citealp{li-etal-2016-diversity}). Table \ref{table_diversity} displays the results of this evaluation. While diversity was slightly reduced in the texts generated through ScoPE during the process of achieving controllability compared to those generated solely by the backbone model, the difference is marginal. Particularly, considering that over 90\% of bigrams and trigrams remain non-duplicative, we can assert that diversity is still well-preserved in the ScoPE generation process.
\begin{table}[t]
    \centering    
    \begin{tabular}{lc}
        \toprule
        \multicolumn{1}{c}{Methods} & Runtime (s) \(\downarrow\) \\
        \midrule
        GPT2-XL & 0.042 \\
        ScoPE ($N$=1) & 0.048 \\
        ScoPE ($N$=5) & 0.050 \\
        ScoPE ($N$=10) & 0.053 \\
        Mix\&Match & 10.826 \\
        \bottomrule
    \end{tabular}
    \caption{Runtime of ScoPE for generation of one token with 10 batch size compared with the GPT2-XL and Mix\&Match baselines}\label{table_runtime}
\end{table}
\begin{table}[t]
\centering
{{
\begin{tabular}{cccc}
\toprule
    \textbf{$b=4$} & \textbf{$b=8$} & \textbf{$b=16$} & \textbf{$b=32$} \\ \hline
     0.057& 0.052 & 0.050 & 0.049 \\
 \bottomrule
\end{tabular}
}}
\caption{ Runtime (seconds) of ScoPE for generation of one token with various block sizes($b$).}
\label{table_block_runtime_abl}
\end{table}
\begin{table*}[t]
\centering
{{
\begin{tabular}{cccccc}
\toprule
    \textbf{Camera} & \textbf{Videogame} & \textbf{Grocery} & \textbf{Music} & \textbf{Pos.} & \textbf{Neg.}\\ \hline
     15.96& 16.35 & 14.32 & 15.98 & 14.72 & 16.03 \\
 \bottomrule
\end{tabular}
}}
\caption{PPL score (measured by LLaMA2-13B) of each Amazon review sub-corpus.}
\label{table_amazon_ppl}
\end{table*}

\section{Analysis of ScoPE} \label{analysis}

\paragraph{Ablation study on the repetition score}
Table \ref{table_anal_abs} demonstrates a significant decrease in the repetition score when it is not utilized during training. In terms of PPL, the repetitive results without repetition score training show improved results. However, this is because language models tend to score higher for repetitive texts. In terms of MAUVE, the presence or absence of the repetition score has a negligible impact on performance for relatively easy in-domain input conditions. However, for out-of-domain conditions, the absence of the repetition score results in a significant performance drop.

\paragraph{Ablation study on the edit block size}
Table \ref{table_anal_block} presents the results of controlled text generation with four different token block sizes, $b=4, 8, 16, 32$. The combined score comprising PPL demonstrates that the small block size setting can guarantee fluency. For out-of-domain input conditions, especially in challenging generation scenarios with lower relevance to the target domain, smaller block sizes show significantly superior MAUVE to larger black sizes. This suggests that reducing the block size for more delicate editing can be beneficial in challenging scenarios.

\paragraph{Analysis of the inference cost of ScoPE}
In the process of generating a total of $T$ tokens, each token undergoing $E$ iterations in the Transformer model, the cost associated with passing a sequence through the model can be denoted as $C$. If we take into account the computational complexity presented by Mix\&Match, it is expressed as $O(T^2EC)$ (\citealp{Goyal2021-rz_mh}). By modifying the computation of the MLM (Masked Language Model) energy calculation, which inherently holds a complexity of $O(TC)$, to be executed in a parallel fashion through a trade-off with memory cost, an optimization towards a time complexity of $O(TEC)$ becomes feasible.

In the scenario of ScoPE, which operates through block-wise edits, the computational complexity is more optimal. During the inference phase, the necessity for MLM energy computation is obviated, and as the generation of $b$ tokens occurs within the context of producing the overall $T$ tokens, the editor conducts $E$ edit operations. Consequently, the computational complexity associated with ScoPE becomes $O(TEC/B)$. This implies that during the token generation process, ScoPE demonstrates advantages over Mix\&Match in both memory and time complexity realms. Table \ref{table_runtime} shows the comparison of the runtime of ScoPE with other baselines.

 During the generation process, the backbone model undergoes a step where it edits the previous text via ScoPE each time $b$ tokens corresponding to the block size are generated. Consequently, there exists a genuine issue of having to recompute hidden states that are usually cached during autoregressive generation each time. However, the cost incurred by this recomputation impacts the overall inference time to a relatively minor degree, as recomputation occurs the times of the $1/b$ of the number of tokens generated. For instance, in the setting of our evaluation with a block size of $b=16$, generating 128 tokens through ScoPE generation, the recomputation of the previous text's hidden states occurs no more than 8 times throughout the entire generation process.
 
Nonetheless, since the frequency of recomputing varies with $b$, it is necessary to measure the inference time for various $b$ values. Therefore, we have measured the ScoPE generation time using LLaMA2-7B as the backbone model across different $b$ values and the results are presented in Table \ref{table_block_runtime_abl}. As expected, a decrease in $b$ leads to an increase in inference time. In extreme cases where $b$ approaches 1 or when the overall length of the generated text becomes very long, a significant issue with inference time may indeed arise.

However, considering that the scope of text edited by ScoPE is limited to the most recently generated $b$ tokens (1) and that tokens generated prior remain unchanged (2), it would be possible to limit the number of tokens whose hidden states need to be recomputed to a maximum of $b$ during the ScoPE generation process. This approach is expected to significantly reduce the cost associated with recomputing hidden states. 

\paragraph{Analysis of the fluency of the text generated by ScoPE}
Despite ScoPE generation achieving improved fluency (as shown by PPL) over existing baselines (DExperts, GeDi, Mix\&Match), we observed an increase in PPL when using ChatGPT and LLaMA2 as backbone models. We interpret this as follows: ScoPE generates text constrained by the overall characteristics of the target corpus including the general fluency of the target corpus.

Table \ref{table_amazon_ppl} presents the average PPL across different subsets of the Amazon review corpus which corresponds to the target corpus on which ScoPE was trained. We note that for category-related sub-corpora, the values are generally around 15, which is consistent with the PPL scores of the text generated through ScoPE as seen in Table \ref{table2}. Additionally, it is observed that for sub-corpora related to sentiment, the Positive corpus generally has a lower PPL than the Negative corpus. This trend also appears consistently in the generation results, where outputs targeting positive sentiment generally have lower PPL than those targeting negative sentiment (as shown in Tables \ref{table5}, \ref{sentiment_llama2_pos}, \ref{sentiment_llama2_neg}). Therefore, while the target score ($s_{\mathrm{mlm}}(\mathbf{x})$) learned by ScoPE includes overall fluency-related information, leading to an improvement in fluency as the number of edit iterations ($N$) increases, the text generated through ScoPE is inherently constrained by the fluency of the target corpus it was trained on.

This interpretation suggests that for the LLaMA2 and ChatGPT backbone models, which generate text with a lower PPL than the original target corpus, it is challenging for ScoPE to produce text with a lower PPL than that of the backbone models. However, as the overall low PPL results in Table \ref{table3} suggest, ScoPE significantly benefits from the fluent text generation capabilities of the backbone model, effectively utilizing its capacity.

\section{Generated Samples}\label{Generated Samples}
We provide examples of the text generated using ScoPE for various controlled text generation tasks in Tables \ref{table_cate_llama} to \ref{table_sent_gpt_samp}.

\section{Usage of AI Writing Assistance} \label{ai_Writing}

This paper was written with linguistic support from the AI assistant ChatGPT, which offered paraphrasing, spell-checking, and polishing of the author's original content. No other assistance was received beyond this support.
\begin{table*}
\centering
{\footnotesize\resizebox{\textwidth}{!}{
\renewcommand{\arraystretch}{1.3}
\begin{tabular}{l||cccccc}
\toprule
\multirow{2}{*}{\textbf{Hyperparams.}} & \multicolumn{6}{c}{\textbf{Domains}} \\ \cline{2-7} 
 & \textbf{Camera} &\textbf{ Videogame} & \textbf{Grocery} & \textbf{Music} & \textbf{Positive} & \textbf{Negative}  \\ \hline
Optimizer & Adam  & Adam & Adam & Adam & Adam & Adam \\
LR peak & 1e-4 & 1e-4 & 1e-4 & 1e-4 &1e-4  &1e-4  \\
LR decaying & Polynomial & Polynomial & Polynomial & Polynomial & Polynomial & Polynomial \\
Weight decay & 0.01 &0.01  &0.01  & 0.01 & 0.01 &  0.01\\
Adam betas & (0.9, 0.98) & (0.9, 0.98) &(0.9, 0.98)  & (0.9, 0.98) &(0.9, 0.98)  & (0.9, 0.98)  \\
Max tokens per batch & 16384 & 16384 & 16384 & 16384 & 16384 & 16384 \\
Update frequency & 2 & 2 & 2 & 2 & 2 & 2 \\
Dropout & 0.1 & 0.1 & 0.1 & 0.1 & 0.1 & 0.1 \\
Attention dropout & 0.1 & 0.1 & 0.1 & 0.1 & 0.1 & 0.1 \\
Max Epoch & 10 & 10 & 10 & 10 & 10 & 10 \\
Warmup ratio & 0.1 &0.1  & 0.1 & 0.1 & 0.1 & 0.1 \\
Training seed & 1 & 1  &1 & 1 & 1 & 1 \\
 \bottomrule
\end{tabular}
}}
\caption{Training hyperparameters for fine-tuning MLM for the target domain. "Update frequency" refers to the number of updates for gradient accumulation.}
\label{table_hyperparam_mlm}
\end{table*}
\begin{table*}
\centering
{\footnotesize\resizebox{\textwidth}{!}{
\renewcommand{\arraystretch}{1.3}
\begin{tabular}{l||cccccc}
\toprule
\multirow{2}{*}{\textbf{Hyperparams.}} & \multicolumn{6}{c}{\textbf{Domains}} \\ \cline{2-7} 
 & \textbf{Camera} &\textbf{ Videogame} & \textbf{Grocery} & \textbf{Music} & \textbf{Positive} & \textbf{Negative}  \\ \hline
Optimizer & Adam  & Adam & Adam & Adam & Adam & Adam \\
LR peak & 1e-4 & 1e-4 & 1e-4 & 1e-4 &1e-4  &1e-4  \\
LR decaying & Polynomial & Polynomial & Polynomial & Polynomial & Polynomial & Polynomial \\
Weight decay & 0.01 &0.01  &0.01  & 0.01 & 0.01 &  0.01\\
Adam betas & (0.9, 0.98) & (0.9, 0.98) &(0.9, 0.98)  & (0.9, 0.98) &(0.9, 0.98)  & (0.9, 0.98)  \\
Max tokens per batch & 2048 & 2048 & 2048 & 2048 & 2048 & 2048 \\
Update frequency & 8 & 8 & 8 & 8 & 8 & 8 \\
Dropout & 0.1 & 0.1 & 0.1 & 0.1 & 0.1 & 0.1 \\
Attention dropout & 0.1 & 0.1 & 0.1 & 0.1 & 0.1 & 0.1 \\
Max Epoch & 10 & 10 & 10 & 10 & 10 & 10 \\
Warmup ratio & 0.1 &0.1  & 0.1 & 0.1 & 0.1 & 0.1 \\
Training seed & 1 & 1  &1 & 1 & 1 & 1 \\
Scale for $s_{\mathrm{rep}}$ & 10.0 & 10.0 & 10.0 & 10.0 & 10.0 & 10.0 \\
Scale for $s_{\mathrm{disc}}$ & N/A & N/A & N/A & N/A & 5.0 & 1.0 \\
 \bottomrule
\end{tabular}
}}
\caption{Training hyperparameters for ScoPE training. "Update frequency" refers to the number of updates for gradient accumulation.}
\label{table_hyperparam_scope}
\end{table*}

\begin{table*}
\centering
{\small
\renewcommand{\arraystretch}{1.2}
\begin{tabular}{p{15cm}}
\toprule
\textbf{System Message}

\texttt{The user did an experiment where several language models were prompted to write a product review. The goal is to write a review with texts of high quality in terms of grammar, spelling, punctuation, word choice, and sentence structure. The user will give you two reviews written by two different models. Briefly explain which review is better and end your answer with: "Preferred writing: 1" or "Preferred writing: 2", depending on which option is better. If both options are equally good, end your response with "Preferred writing: None".}
\\ \hline
\textbf{User Message}

\texttt{[Model 1]}

\{\texttt{TEXT GENERATED FROM MODEL 1}\}

\texttt{[Model 2]}

\{\texttt{TEXT GENERATED FROM MODEL 2}\}
\\ \bottomrule
\end{tabular}
}
\caption{Prompts content used for the GPT-4 evaluation for fluency. The prompt content refers to \citet{controlled_arithmetic}.}
\label{table_gpt4eval_prompt_fluency}
\end{table*}
\begin{table*}
\centering
{\small
\renewcommand{\arraystretch}{1.2}
\begin{tabular}{p{15cm}}
\toprule
\textbf{System Message}

\texttt{The user did an experiment where several language models were prompted to write a product review. The goal is to write a review that is relevant to the camera products. The user will give you two reviews written by two different models. Briefly explain which review is better and end your answer with: "Preferred writing: 1" or "Preferred writing: 2", depending on which option is better. If both options are equally good, end your response with "Preferred writing: None".}
\\ \hline
\textbf{User Message}

\texttt{[Model 1]}

\{\texttt{TEXT GENERATED FROM MODEL 1}\}

\texttt{[Model 2]}

\{\texttt{TEXT GENERATED FROM MODEL 2}\}
\\ \bottomrule
\end{tabular}
}
\caption{Prompts content used for the GPT-4 evaluation for controllability. The prompt content refers to \citet{controlled_arithmetic}.}
\label{table_gpt4eval_prompt_cont}
\end{table*}

\begin{table*}
\centering
{\small
\renewcommand{\arraystretch}{1.2}
\begin{tabular}{p{15cm}}
\toprule
\textbf{System Message}

\texttt{You are proficient assistant for the text completion task. Generate the continuation for the given text. Generate at least 128 tokens. Additionally, the continuation must become Amazon review of cameras like followings: \{EXAMPLES FROM TARGET CORPUS\}}
\\ \hline
\textbf{User Message}

\texttt{[Text]}

\texttt{This text is about}

\texttt{[Continuation]}
\\ \bottomrule
\end{tabular}
}
\caption{Prompts content used for the ChatGPT instruction for controllability.}
\label{table_chatgpt_prompt_cont}
\end{table*}

\begin{table*}
\centering
\footnotesize\resizebox{\textwidth}{!}{
\renewcommand{\arraystretch}{1.3}
\begin{tabular}{l|c||cccc|cccc}
\toprule
\multirow{2}{*}{\textbf{Target Attributes}} & \multirow{2}{*}{\textbf{Methods}} & \multicolumn{4}{c|}{$\textbf{PPL}$ $\downarrow$} & \multicolumn{4}{c}{\textbf{MAUVE} $\uparrow$} \\
& & \textbf{Camera} & \textbf{Videogame} & \textbf{Grocery} & \textbf{Music} & \textbf{Camera} &\textbf{Videogame} & \textbf{Grocery} & \textbf{Music} \\ \hline
\multirow{4}{*}{\textbf{Camera}}& LLaMA2-7B  & 11.9& 12.34& 10.51& 12.06
& 0.6643& 0.1434& 0.0654& 0.0290\\
& ScoPE ($N$=1) & 13.06&  14.73 & 14.22 & 16.51
& 0.7159& 0.5658& 0.4099& 0.3030\\
& ScoPE ($N$=5) & 11.47&  12.67&  12.38&  14.04
& 0.7415& 0.6621& 0.5439& 0.4740\\
& ScoPE ($N$=10) & 11.46 & 12.52&  12.12& 13.96& 0.7600& 0.6654& 0.5221& 0.4758\\ \hline
\multirow{4}{*}{\textbf{Videogame}} & LLaMA2-7B  & 11.9& 12.34& 10.51& 12.06
& 0.1738& 0.7023& 0.0752& 0.0807
\\
& ScoPE ($N$=1) & 14.9& 13.56& 13.96& 15.19
& 0.4991& 0.7079& 0.3283& 0.4095
\\
& ScoPE ($N$=5) & 12.65& 11.72& 12.04& 13.12
& 0.562& 0.7329& 0.4175& 0.4851
\\
& ScoPE ($N$=10) & 12.47& 11.66& 12.02& 13.03& 0.5779& 0.7286& 0.443& 0.5039\\ \hline
\multirow{4}{*}{\textbf{Grocery}} & LLaMA2-7B & 11.9& 12.34& 10.51& 12.06
& 0.0467& 0.0397& 0.4952& 0.0265
\\
& ScoPE ($N$=1) & 14.35& 14.44& 11.73& 15.46
& 0.1552& 0.1508& 0.584& 0.1342
\\
& ScoPE ($N$=5) & 12.11& 12.19& 10.17& 13
& 0.2858& 0.289& 0.6364& 0.2468
\\
& ScoPE ($N$=10) & 11.86& 12.18& 10.2& 12.98& 0.2926& 0.3279& 0.6227& 0.2715\\  \hline
\multirow{4}{*}{\textbf{Music}} & LLaMA2-7B & 11.9& 12.34& 10.51& 12.06
& 0.0395& 0.0798& 0.0462& 0.7618
\\
& ScoPE ($N$=1) & 16.42& 15.6& 15.23& 14.51
& 0.1718& 0.2483& 0.2091& 0.6354
\\
& ScoPE ($N$=5) & 14.4& 13.96& 13.67& 13.18
& 0.2703& 0.3353& 0.2808& 0.6265
\\
& ScoPE ($N$=10) & 14.35& 13.91& 13.56& 13.32& 0.2646& 0.3431& 0.2869& 0.6204\\
\bottomrule
\end{tabular}
}
\caption{Experimental results for category controlled generation targeting Camera, Videogame, Grocery, and Music attributes. ScoPE utilizes LLaMA2-7B for the backbone model as black-box.}
\label{category_llama2_others}
\end{table*}
\begin{table*}
\centering
{\footnotesize\resizebox{\textwidth}{!}{
\renewcommand{\arraystretch}{1.3}
\begin{tabular}{l||cccc|cccc}
\toprule
\multirow{2}{*}{\textbf{Methods}} & \multicolumn{4}{c|}{$\textbf{PPL}$ $\downarrow$} & \multicolumn{4}{c}{\textbf{MAUVE} $\uparrow$} \\
  & \textbf{Camera} & \textbf{Videogame} & \textbf{Grocery} & \textbf{Music} & \textbf{Camera} & \textbf{Videogame} & \textbf{Grocery} & \textbf{Music}  \\ \hline
 ScoPE ($N$=1) &         13.17& 14.95& 14.26& 16.29& 0.8511& 0.687& 0.607& 0.4403
\\ 
  no $\mathbf{s_{\mathrm{rep}}}$  &       12.08& 12.91& 14.61& 13.54& 0.8555& 0.5632& 0.4778& 0.2644
\\
 ScoPE ($N$=5) &         11.86& 13.22& 12.73& 14.39& 0.8366& 0.7396& 0.6594& 0.5757
\\ 
  no $\mathbf{s_{\mathrm{rep}}}$  &       10.96& 11.75& 13.05& 11.86& 0.8315& 0.6427& 0.5968& 0.3811
\\ 
  ScoPE ($N$=10) &        11.73& 12.96& 12.69& 14.37& 0.8266& 0.7441& 0.6815& 0.5788
\\ 
 no $\mathbf{s_{\mathrm{rep}}}$  &        10.77& 11.64& 12.88& 11.66& 0.836& 0.6472& 0.6014& 0.4262\\
\bottomrule
\end{tabular}
}}
\caption{Ablation studies about repetition score for category controlled generation targeting the camera domain. "no $\mathbf{s_{\mathrm{rep}}}$" denotes not using repetition score for ScoPE training.}
\label{table_anal_abs}
\end{table*}
\begin{table*}
\centering
{\footnotesize\resizebox{\textwidth}{!}{
\renewcommand{\arraystretch}{1.3}
\begin{tabular}{l|l|cccc|cccc}
\toprule
\multirow{2}{*}{\textbf{Block Size}} & \multirow{2}{*}{\textbf{Methods}} & \multicolumn{4}{c|}{{$\textbf{PPL}$ $\downarrow$}} & \multicolumn{4}{c}{\textbf{MAUVE} $\uparrow$} \\
& & \textbf{Camera} & \textbf{Videogame} & \textbf{Grocery} & \textbf{Music} & \textbf{Camera} &\textbf{Videogame} & \textbf{Grocery} & \textbf{Music}   \\ \hline
\multirow{3}{*}{4} &  ScoPE ($N$=1)             &  12.79& 13.85&  13.54&   15.32
&  0.7884&  0.73&   0.6038&  0.6055
\\
&   ScoPE ($N$=5)                &   11.8&  12.61&    12.43&  13.73
&  0.7825&    0.7376&  0.6537& 0.6477
\\
&   ScoPE ($N$=10)                &  11.68&   12.53&  12.28&   13.67&   0.7754&  0.7364&  0.6227& 0.6363
\\ \hline
\multirow{3}{*}{8} &   ScoPE ($N$=1)                &   12.71&  14.06&   13.76&    15.45
&   0.8313&     0.7359&   0.6613&   0.5475
\\
&   ScoPE ($N$=5)                &  11.78&   12.69&     12.36&   13.87
&  0.8197&    0.7588&   0.6613&   0.6689
\\
&   ScoPE ($N$=10)                &   11.6&  12.47&  12.26&   13.64&  0.8302&  0.7574&  0.6743&  0.6558
\\ \hline
\multirow{3}{*}{16} &  ScoPE ($N$=1)             &  13.17& 14.95&  14.26&   16.29
&  0.8511&  0.687&   0.607&  0.4403
\\
&   ScoPE ($N$=5)                &   11.86&  13.22&    12.73&  14.39
&  0.8366&    0.7396&  0.6594& 0.5757
\\
&   ScoPE ($N$=10)                &  11.73&   12.96&  12.69&   14.37&   0.8266&  0.7441&  0.6815& 0.5788
\\ \hline
\multirow{3}{*}{32} &   ScoPE ($N$=1)                &   13.39&  15.12&   14.87&    16.46
&   0.8418&     0.5987&   0.4542&   0.2743
\\
&   ScoPE ($N$=5)                &  12.21&   13.26&     13.09&   15.01
&  0.8497&    0.7178&   0.6287&   0.4645
\\
&   ScoPE ($N$=10)                &   12.03&  13.17&  12.91&   14.85&  0.8412&  0.7201&  0.6481&  0.4759\\ 
\bottomrule
\end{tabular}
}}
\caption{Analysis about the impact of various token block sizes for ScoPE editing.}
\label{table_anal_block}
\end{table*}

\begin{table*}
\centering
{\small
\renewcommand{\arraystretch}{1.2}
\begin{tabular}{p{2cm} || p{13cm}}
\toprule
\textbf{Input Text} & This text is about
\\ \hline
\textbf{Camera Cont.} &
This text is about  2 years old, but it still has good quality.
The size is perfect to be carried in your purse and just take it anywhere. :)

The purpose of this product is
to provide a simple and easily-accessible solution for the owner of a digital camera/lENS who want an inexpensive solution that gives full control over their camera settings, such as shutter speed, aperture and ISO settings.

The only problem with this particular Canon lens hood for Canon cameras is the size of your lens. This is because your lens fits in the hood and it prevents dust from getting into the front of your lens, which can potentially cause
\\ \hline
\textbf{Videogame Cont.} & 
This text is about  2 years old, but it still has great graphics.
This game is great, my friend bought it for his Atari 6600, and we played it as much as we used to play games over 20 years ago. We had a blast

This game is great for two players, because you can play the 1st player with A or B, and the second with C or D.
Buy from[...] (http://amazon.com)

Downloadable version of atarimania[...] (http://atarimania.com) 
Some information in the help file:
http://atarimania.
\\ \hline
\textbf{Grocery Cont} & 
This text is about  20 years old, but it still has great flavor.
This product is great to be used in combination with the other 'Bob Red Mill' products, but it stands alone very well. I prefer the texture and taste of all the Bob products. If you're wondering what that all means, this is a good product.
Ingredients: Sprouted Whole Grain Wheat, Virgin Olive Oil, Expe34;34; Mix this product with 2 cups of water and 2 tablespoons milk. Mix and simmer for 6 hours before using. Use it every day!
Ingredients for the Organic Oatmeal Oatmeal: Sprouted Whole Grain Wheat,
\\ \hline
\textbf{Music Cont.} & 
This text is about  20 years old, but it still has great meaning.
The text is available at my website www.all-music.com, or my website \"all-music,\" and the \"Reunion\" Album is an anthology of songs from my 20 year career in music, and it is at ALLBABY.com. I also have a new Christmas CD called:\"MERRY HOME\" which has some great Christmas songs. This new Christmas CD has two new Christmas songs and two traditional Christmas songs. It is a very good new Christmas CD and I think you will enjoy it. It was new to
\\
 \bottomrule
\end{tabular}
}
\caption{Generated samples from the category controlled generation whose backbone language model is LLaMA2-7B.}
\label{table_cate_llama}
\end{table*}
\begin{table*}
\centering
{\small
\renewcommand{\arraystretch}{1.2}
\begin{tabular}{p{2cm} || p{13cm}}
\toprule
\textbf{Input Text} & This text is about
\\ \hline
\textbf{Camera Cont.} &
This text is about the use and understanding of techniques and methods for producing goods and services. The paper defines photography and video work as a creative endeavor. There is no limit, there is an upper limit and there is an lower limit. 
Some things to note in the B\&H Photo review of this product is that the author says that the Hoya Super HMC UV (Black) filter is coated on both sides, with an additional coating on the inside (similar to that of a clear lens cap), and says that it is designed for UV use.  This is a false statement.  Hoya is apologizing for this by not unsubscribing
\\ \hline
\textbf{Videogame Cont.} & 
This text is about the legal rules for this type of game.

Self-expression is a powerful force for good (or bad),
influencing attitudes and influencing behavior between a large number of people. 
The self-expression is very powerful, and that means I love this game. 
Attitude is very important part of life. If your attitude is good, you will make yourself better. 
If you have bad attitude, you won’t like this game anyway.
The best thing about this game is its graphics. After playing this game, you can tell that I am happy. 
The overall quality of
\\ \hline
\textbf{Grocery Cont} & 
This text is about the quality of coffee at colleges and universities, and is a revised version of the author's original written review: http://digitalcommons.chapel.food@colorado.edu food@colorado.edu University of Florida coffee@Florida.edu  email (at)metrofood.com  http://www.thefoodnetwork.com There are a few products and services available on this site: food@foodnetwork.com  http://www.traderfood.com www.coffee-company.com  focus on your health and understand the importance of doing it right http://www.detroit
\\ \hline
\textbf{Music Cont.} & 
This text is about people who will pay for you if you want to try and beat them.

 My MP3 player is a CD and over the last decade these songs were all good but they do not mean much to me anymore. \"Stuck in a Place You Can’t Get Out of\" is not new to me and that’s probably why it doesn’t rank well for me anymore, which sucks because I think it’s such a good song.  I am going to review this album in a way that I’m sure makes some people cringe, but what can I say I’m
\\
 \bottomrule
\end{tabular}
}
\caption{Generated samples from the category controlled generation whose backbone language model is \texttt{davinci-002}.}
\label{table_cate_davinci}
\end{table*}
\begin{table*}
\centering
{\small
\renewcommand{\arraystretch}{1.2}
\begin{tabular}{p{2cm} || p{13cm}}
\toprule
\textbf{Input Text} & This text is about
\\ \hline
\textbf{Camera Cont.} &
This text is about learning one’s photography.
Thanks.
Now I think I am ready to take pictures

Watched a video here on how to read the histogram and such. It was very helpful. So much to learn about it! And how everything about it affects your pictures. If you pay attention to the buttons that you press on your camera you can see what they are doing, and that's what I did. Press one of them, hold it down and then take pictures. Then adjust the white balance and image quality for portrait or landscape.  Done! 
Also, when you go into the menu after taking pictures
\\ \hline
\textbf{Videogame Cont.} & 
This text is about playing a semi-fun, madd’em on-line game, multi-ball golf. Just like all the pachinko games in Japan this type of game is fun and filled with jolly music, good graphics and new players come bursting out of the screen.

If you are a fan of Halo and want to read a “review” of one of the previous games, there are many reviews for this game coming out in March. I hope you enjoy!

Halo 3 is really fun. It ends right where the first one ended and it just falls right into
\\ \hline
\textbf{Grocery Cont} & 
This text is about the video player.  In other words, avoid this product. The description is woefully misleading and I will not buy this product ever again. I love it is the easiest way to play video and it has good price as well.

 As a coffee lover this allows me to listen to my coffee when I’m trying to fall asleep and my kids are reading a book on the computer. It’s very convenient as well!  This is a product that tastes good, smells good, and you can feel good about giving to your kids and yourself. Plus, this stuff is not only good for you, but it tastes
\\ \hline
\textbf{Music Cont.} & 
This text is about the music, except for Rock \& Roll and other genres. See my review for Superhighway for information about the music:\"Before I die, I would love for this music to be on the radio\".  Art Tatum was popular in an era of extrovert showmanship. His audience loved him. It was also an era of unemployment when a lot of people were out of work. They wanted to hear something and enjoy, not sit back and listen. This is a collection of his songs. Some people say Frankie Laine influenced Art Tatum in that he could be heard up close and personal. Others say he
\\
 \bottomrule
\end{tabular}
}
\caption{Generated samples from the category controlled generation whose backbone language model is \texttt{babbage-002}.}
\label{table_cate_babbage}
\end{table*}
\begin{table*}
\centering
{\small
\renewcommand{\arraystretch}{1.3}
\begin{tabular}{p{2cm} || p{13cm}}
\toprule
\textbf{Input Text} & "role": "system", "content": You are proficient assistant for the text completion task. Generate the continuation for the given text. Generate at least 128 tokens.

"role": "user", "content": [Text] This text is about [Continuation] 

\\ \hline


\textbf{Baseline Cont.} &  
This text is about the importance of self-care and maintaining a healthy lifestyle. In today's fast-paced world, it is easy to get caught up in the hustle and bustle of our daily lives, often neglecting our own well-being in the process. However, taking the time to prioritize self-care is crucial for our overall physical, mental, and emotional health.

Self-care can take many forms, and it looks different for everyone. It may involve carving out time for relaxation and taking part in activities that bring us joy and rejuvenation. This could include practicing yoga, meditating, reading a book, or indulging
\\ \hline
\textbf{Baseline} +

\textbf{Instruction Cont.}
 &  
 This text is about a camera I recently purchased from Amazon. I have to say, I am extremely disappointed with the quality of this camera. The battery compartment is a major issue, just like many other reviewers have mentioned. The battery simply won't stay latched in place, causing much frustration during photo sessions. However, I did come up with a temporary fix to this problem. I folded a small post-it note twice and jammed it between the battery and the grip, and miraculously, it keeps the battery jammed in place. It's definitely a lo-fi fix, but it gets the job done.

Aside from the battery issue, the
 \\ \hline
\textbf{ScoPE Cont.}
 & 
 This text is about:
The importance of photography and how it affects individuals' lives. It does not only capture moments and preserve memories, but also serves as a great form of self-expression and creativity. Through photography, individuals can express their emotions and thoughts in ways that they cannot otherwise do. It allows them to capture a special moment in life and turn it into a beautiful piece of art. And it's simply a way to preserve memories and provide a tangible document of one's life. Photography also opens up new opportunities and creative options for individuals. With the advent of social media sites, individuals can easily share their pictures with a wider audience and get positive
 \\ \hline
\textbf{ScoPE} +

\textbf{Instruction Cont.}
 & 
 This text is about the camera I just purchased. Same problem as other people have had: the lens won't stay latched on. So, same problem as others with the lens not staying latched on, but I have found a simple solution: I attached a small piece of velcro to the lens and now it stays securely in place. It's a shame that a simple thing like keeping the lens latched on properly is an issue with this camera. Also, the picture quality is not as good as I expected. The pictures come out blurry and lack sharpness, even in well-lit areas. The autof
 \\
 \bottomrule
\end{tabular}
}
\caption{Generated samples from the category controlled generation targeting camera attributes whose backbone language model is \texttt{gpt-3.5-turbo-0613}.}
\label{table_chatgpt_samp}
\end{table*}
\begin{table*}
\centering
{\small
\renewcommand{\arraystretch}{1.2}
\begin{tabular}{p{2cm} || p{13cm}}
\toprule
\textbf{Input Text} & This text is about
\\ \hline
\textbf{Pos. Cont.} &
This text is about  50 years old, but it still has great quality.
This book is great to read if you're into the "dark" (i.e., ). I think this book is one of the best, and if not one of the best, in the very "dark" genre.

By John J. O'Brien. Thanks

Cover art is from the original CD P\&S/CD \#1 (2).

Thank you!
Thank you so much!
\\ \hline
\textbf{Pos. Cont.}

($+$ EDG) & 
This text is about  5 years old, but it still has great graphics.
This game is great, my friends and I used to play it all the time (and still do today).
We had lots of fun with this game, and last year we both bought some new computers and we played some of the old games we used to play with much bigger and better graphics. We still love this game, you can definitely buy it. Only complaint is the price. I looked online and it only was \$10 but I think that's too high because I paid \$15 right when I bought it at the store (5 yrs ago). Overall a great
\\ \hline
\textbf{Neg. Cont.} & 
This text is about  2 years old, but it still has good graphics

This game is not like the previous games. It has less "depth" and story. 
"The idea here is: "You have a space ship."
But I don't really have any idea of how to play it, or even the point. 
There may be something better out there here, but it's hard to find it because the game-developer hasn't released it.
The game has a lot of potential; however, it's been out for a while. 
And I'm sure most people have moved on.

Bad
\\ \hline
\textbf{Neg. Cont.}

($+$ EDG) & 
This text is about  2 years old, but it still has good graphics.
This game is not like the previous games. It has no "story" (and it's kind of boring).
You can only play 1 or 2 hours (it goes on for an hour).
And all the games are on a small screen, so you see much less than in the previous games.
You can't turn off the camera to take a picture of the game on your computer (because it's the same as a camera).
If you play on your computer, you get a lot of bugs in this game (the graphics are terrible).
You have
\\
 \bottomrule
\end{tabular}
}
\caption{Generated samples from the sentiment controlled generation whose backbone language model is GPT2-XL.}
\label{table_sent_gpt_samp}
\end{table*}

\end{document}